\documentclass[twoside]{article}
\usepackage{arxiv,palatino,epsfig,latexsym,natbib}

\usepackage{multirow}
\usepackage[normalem]{ulem}
\usepackage[T1]{fontenc}

\usepackage{lipsum}

\usepackage{algorithm,algorithmicx,algpseudocode}

\usepackage{amssymb}
\usepackage{amsmath}

\usepackage{url}

\usepackage[scientific-notation=true]{siunitx}

\usepackage{xargs}
\usepackage[pdftex,dvipsnames,table]{xcolor}  
\usepackage[colorinlistoftodos,prependcaption,textsize=normal]{todonotes}

\newcommandx{\unsure}[2][1=]{\todo[linecolor=red,backgroundcolor=red!25,bordercolor=red,#1]{#2}}
\newcommandx{\change}[2][1=]{\todo[linecolor=blue,backgroundcolor=blue!25,bordercolor=blue,#1]{#2}}
\newcommandx{\info}[2][1=]{\todo[linecolor=OliveGreen,backgroundcolor=OliveGreen!25,bordercolor=OliveGreen,#1]{#2}}
\newcommandx{\improvement}[2][1=]{\todo[linecolor=Plum,backgroundcolor=Plum!25,bordercolor=Plum,#1]{#2}}
\newcommandx{\thiswillnotshow}[2][1=]{\todo[disable,#1]{#2}}

\begin{document}

\ecjHeader{x}{x}{xxx-xxx}{201X}{Data-Efficient Design Exploration through Surrogate-Assisted Illumination}{A Gaier, A Asteroth, J-B Mouret}
\title{\bf Data-Efficient Design Exploration through Surrogate-Assisted Illumination}  

\author{\name{\bf Adam Gaier} \hfill 
		\addr{adam.gaier@h-brs.de}\\ 
    \addr{Universit\'e de Lorraine, CNRS, Inria, LORIA, F-54000 Nancy, France}\\
    \addr{Bonn-Rhein-Sieg University of Applied Sciences, Sankt Augustin, 53757, Germany}
\AND
       \name{\bf Alexander Asteroth} \hfill 
       \addr{alexander.asteroth@h-brs.de}\\
        \addr{Bonn-Rhein-Sieg University of Applied Sciences, Sankt Augustin, 53757, Germany}
\AND
       \name{\bf Jean-Baptiste Mouret} \hfill \addr{jean-baptiste.mouret@inria.fr}\\
       \addr{Universit\'e de Lorraine, CNRS, Inria, LORIA, F-54000 Nancy, France}
}

\maketitle

\begin{abstract}

Design optimization techniques are often used at the beginning of the design process to explore the space of possible designs. In these domains illumination algorithms, such as MAP-Elites, are promising alternatives to classic optimization algorithms because they produce diverse, high-quality solutions in a single run, instead of only a single near-optimal solution. Unfortunately, these algorithms currently require a large number of function evaluations, limiting their applicability. In this article we introduce a new illumination algorithm, Surrogate-Assisted Illumination (SAIL), that leverages surrogate modeling techniques to create a map of the design space according to user-defined features while minimizing the number of fitness evaluations. On a 2-dimensional airfoil optimization problem SAIL produces hundreds of diverse but high-performing designs with several orders of magnitude fewer evaluations than MAP-Elites or CMA-ES. We demonstrate that SAIL is also capable of producing maps of high-performing designs in realistic 3-dimensional aerodynamic tasks with an accurate flow simulation. Data-efficient design exploration with SAIL can help designers understand what is possible, beyond what is optimal, by considering more than pure objective-based optimization.
\end{abstract}




\section{Introduction}


Creators of design optimization techniques often think of their algorithms as a finalizing step in the design process. Imagining that their techniques will be used to push the limits of performance, they judge success by the ability of an algorithm to refine a design to its most optimal form~\citep{thompson1996evolved,renner2003genetic,hornby2011computer}, with the ultimate goal of outperforming the best engineers.


If, however, the goal of these algorithms is to support designers in discovering the best possible designs, the emphasis on optimal performance may be misplaced. In an interview study by Autodesk~\citep{Bradner2014}, it was found that computational design tools
most common use was \textit{not} at the end of the design process to hone existing designs. Instead the engineers, designers, and architects interviewed reported that they more commonly used optimization tools at the beginning of the design process to explore the space of possible designs. 
By using optimization tools to produce a range of design alternatives, designers explore differing design concepts, allowing them to examine the trade-offs each alternative represents. The designs generated are a consequence of the problem definition, solution constraints, and the assumptions inherent in the design's representation. Once these factors are reconsidered and adjusted, the optimization algorithm is run again to generate new designs, and the process is repeated.

Designers use this generative design cycle to explore and describe complex design spaces. By taking high-performing solutions as concrete way points, they develop a vocabulary to better describe and understand design spaces that are often rendered opaque by their complexity. Armed with this understanding, designers work within these known high-performing design regions, refining designs in light of considerations which are difficult to formalize, including intangibles such as aesthetics.


The most commonly used method of producing this variety of high performing designs is multi-objective optimization~\citep{Deb2003,deb2006innovization}. Instead of searching for a single solution multi-objective optimization produces a Pareto front of non-dominated solutions and, when the objectives are in conflict, each design represents a trade-off between the objectives~\citep{Deb2003}. During the explorative process, however, interest for designers often lies not only in the optimization of all objectives, but in the effect of different design features on performance. Exploration with multi-objective approaches becomes particularly problematic when the objectives are \emph{not} in conflict, and only a few variations of a dominant design theme results. In addition, focusing on pure optimization may lead to solutions which ``overfit'' the objectives, whereas a designer could use expert knowledge to recognize that a sub-optimal design -- according to the objectives -- is actually better suited for the application.

To probe the search space for interesting designs and design principles, new algorithms created specifically for exploration, known as quality-diversity algorithms, could be applied \citep{Lehman2011,lehman2011abandoning,Mouret2015,Pugh2016}. One such algorithm, MAP-Elites~\citep{Mouret2015}, explicitly explores the relationship between user-defined features and performance to produce low-dimensional ``maps'': given features deemed interesting or important, such as weight or structural strength, MAP-Elites produces a large set of high-performing solutions which span the possible variations of those features. 
This \emph{illumination} process reveals the performance potential of the features in varying degrees and combinations.

While MAP-Elites is effective at finding diverse high-performing solutions, the search process requires a tremendous number of evaluations. The illumination process which produced the repertoire of hexapod controllers in~\cite{cully2015robots}, for example, required twenty million evaluations. In applications such as structural optimization or fluid dynamics, where a single evaluation can take hours, this is simply infeasible.

In computationally expensive problems it is common to use surrogate models, that is, approximate models of the objective function based on previously evaluated solutions~\citep{Jin2005, Forrester2009, Shahriari2016, Preen2016design}.
These models are refined through iterative evaluation of new solutions based on an \emph{acquisition function}, which balances exploitation and exploration to improve accuracy in high-fitness regions.
By substituting expensive objective functions with these computationally efficient approximations, the optimization process can be greatly accelerated. 

\begin{figure}[ht]
	\includegraphics{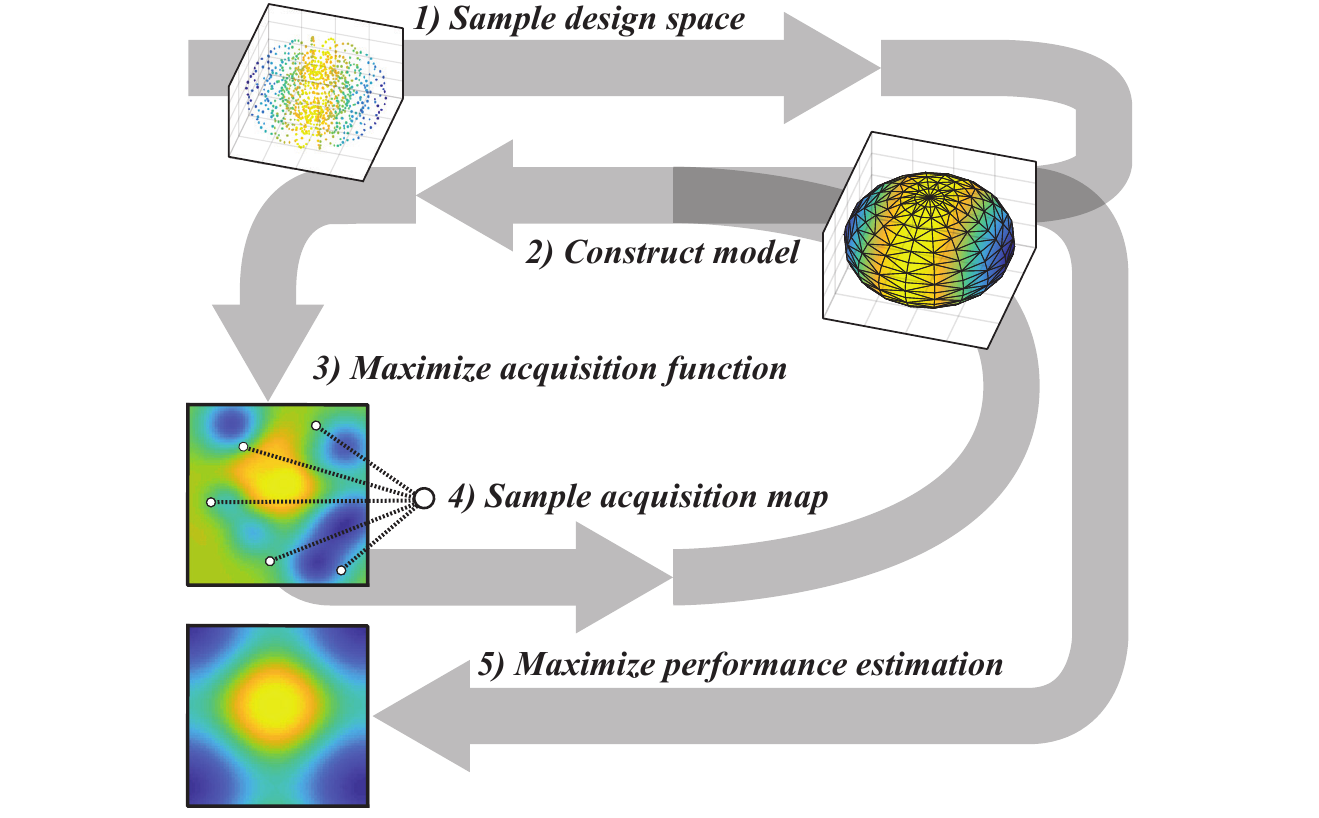}
	\centering
	\caption[caption]
	{	
		\textit{Surrogate-Assisted Illumination (SAIL)} \\\hspace{\textwidth} 
		\textit{1)} Sample design space to produce initial solutions.
		\textit{2)} Construct model of objective function based on samples.
		\textit{3)} Maximize the acquisition function, balancing exploitation and exploration, in every bin of the feature space with MAP-Elites, producing an \emph{acquisition map}.
		\textit{4)} Draw samples from the \emph{acquisition map} to test on the objective function. 
		Repeat steps 2-4 as computational budget allows.
		\textit{5)} Maximize fitness, as predicted by the resulting model, to produce a \emph{prediction map} with MAP-Elites. 
	}
	\label{fig:alg_overview}
\end{figure}

The key insight leveraged in this work is that surrogate models can not only lessen the burden of \emph{expensive} evaluations, but also \emph{numerous} evaluations. 
Incorporating surrogate-assistance techniques into the evaluation-heavy illumination process has the potential to make MAP-Elites efficient enough for use in computationally expensive problems, and accelerate it in even inexpensive problems by orders of magnitude.

In this article, we present the Surrogate-Assisted Illumination (SAIL) algorithm to improve the efficiency, and so expand the applicability, of MAP-Elites. The value of integrating surrogate models into illumination relies on reducing computational cost while maintaining MAP-Elites' original capabilities, resulting in an algorithm that is:  
\begin{itemize}
	\itemsep-0.25em
	\item \textit{Divergent} - Produces a solutions which vary across a user-defined continuum;
	\item \textit{Accurate} - Predicts behavior of the objective function in high-performing regions;	
	\item \textit{High-Performing} - Produces near-optimal solutions;
	\item \textit{Data-Efficient} - Performs even when functions evaluations are expensive or limited.
\end{itemize}

SAIL is a Bayesian optimization corollary for illumination, leveraging modeling techniques to accelerate MAP-Elites. In general terms this interaction between illumination and modeling proceeds as follows (Figure~\ref{fig:alg_overview}):
a surrogate model is constructed based on a set of initial solutions,  
MAP-Elites is used to produce solutions that maximize the acquisition function in every region of feature space, yielding an \emph{acquisition map}, new samples are then drawn from the acquisition map and evaluated, and these additional observations are used to improve the model. 
This acquisition process is repeated to produce increasingly accurate models of the high-fitness regions of the feature space. 
Performance predictions of the model can then be used by MAP-Elites in place of the objective function to produce a \emph{prediction map} of estimated optimal designs in all feature regions.


\subsection*{Notes on Previous Work}
This article is an extended version of `Data-Efficient Exploration, Optimization, and Modeling of Diverse Designs through Surrogate-Assisted Illumination' \citep{gaier2017data}.
In that preliminary work, SAIL was introduced and evaluated in a 2D airfoil domain to demonstrate its ability to produce a diversity of high-performing designs through the data-efficient creation of surrogate models that span the chosen feature dimensions. These results are summarized in Section \ref{sec:2d}. 

In~\cite{gaier2017data}, several assumptions and simplifications were made in order for experiments and analysis to remain tractable. The present work demonstrates that SAIL can work in more realistic settings. Firstly, though SAIL is presented as a data-efficient method prepared for use in expensive problems, the two-dimensional airfoil problem is, with modern computers, not truly expensive. While this allowed for comparison to less data-efficient approaches, along with exhaustive evaluation of solutions throughout the process, it did not demonstrate SAIL's suitability for use in computationally expensive domains. Here we apply SAIL to a three-dimensional aerodynamics case using a fully featured computational fluid dynamics simulator.

Secondly, individual parameters were chosen as features dimensions to allow for straight-forward comparison with standard black-box optimization approaches which do not consider features. In the airfoil case these parameter values are also design features, but this choice nonetheless obscures the purpose of SAIL: to explore user-chosen features dimensions, regardless of the solution representation. In this work, we define feature dimensions unaligned with parameter values, demonstrating that SAIL is doing more than parallel search in a partitioned parameter space. 

Finally, the encoding used in the two-dimensional airfoil experiment is itself the result of extensive analysis and experimentation in a heavily researched domain, built for optimization success~\citep{Sobieczky1999}. This powerful representation provided us with parameterized features well suited for optimization and modeling. To tackle a new non-standard domain we use an untested purpose-built representation as well as a general deformation approach, demonstrating SAIL's ability to effectively explore a low-dimensional feature space regardless of the power of the underlying encoding.

 \section{Related Work} \label{sec:relWork}
 	\subsection*{Quality Diversity and MAP-Elites}\label{sec:qd}

Quality diversity (QD) algorithms~\citep{Pugh2016, Mouret2015, cully2015robots} use evolutionary methods to produce a set of diverse, high-quality solutions within a single run. 
Rather than seeking a single global optimum, the goal of QD algorithms is to discover as many different types of solutions to a problem as possible, and produce the best possible example of each type. 
For this reason they are also referred to as \emph{illumination} algorithms, as they illuminate the performance potential of different regions of the solution space. Two core approaches have emerged to produce this illumination: Novelty Search with Local Competition (NSLC), and Multi-dimensional Archive of Phenotypic Elites (MAP-Elites).

NSLC~\citep{Lehman2011} uses a multiobjective approach to combine rewards for performance and novelty. 
The population is divided into niches based on phenotypic similarity (e.g. behavior or features) and the individual's performance is judged locally against other members of their niche. Novelty is judged globally, with individuals rewarded based on dissimilarity to their neighbors. In this way exploration of the search space and exploitation of existing niches is simultaneously pursued.

The  MAP-Elites algorithm~\citep{Mouret2015, cully2015robots} is designed to produce high-performing solutions across a continuum of $n$ user-defined feature dimensions. It first divides the feature space into an n-dimensional grid, or map, of bins, with one dimension for each feature. The bins are typically squares, but it is also possible to define bins of more complex shapes to control their number precisely \citep{vassiliades2017using}. The map houses the population of solutions, with each bin holding a single solution. When the map is visualized, with each bin colored according to the performance of the solution it contains, it provides an intuitive overview of the performance potential of each region of the feature space.

MAP-Elites begins with the creation of a set of random solutions which are evaluated and assigned to bins according to their features.  If, for example, the feature space has one dimension for weight and another for cost, a low-cost and low-weight solution would be placed in the low-cost, low-weight bin of the map. If the bin is empty, the solution is placed inside. If there is already a solution in the bin then the two solutions are compared, with the higher fitness solution earning or retaining its place in the bin and the lower fitness solution discarded. As a result of these comparisons, each bin contains the best solution found so far for each combination of features. These solutions are known as \emph{elites}.

To produce new solutions, parents are chosen randomly from the elites, mutated, evaluated, and assigned a bin based on their features. Child solutions have two ways of joining the breeding pool: discovering an unoccupied bin, or out-competing an existing solution for its bin. By repeating this simple process of selection, recombination, and bin assignment, a set of increasingly optimal solutions is produced and the feature space increasingly explored, \emph{illuminating} the performance potential of every region of the feature space. MAP-Elites is summarized in Algorithm \ref{alg:mapelites}.

\begin{algorithm} [h!]
  \caption{Multi-dimensional Archive of Phenotypic Elites (MAP-Elites)}
  \label{alg:mapelites}
  \begin{algorithmic}[1]
    \Function{MAP-Elites}{$objective\_function()$, $\mathcal{X}_{initial}$}
    \State $\mathcal{X} \gets \emptyset$,  $\mathcal{P} \gets \emptyset$ 
    \Comment{\textit{Create empty map for genomes $\mathcal{X}$, and performances $\mathcal{P}$}}
   
    \State $\mathcal{X} \gets \mathcal{X}_{initial}$
    \Comment{\textit{Place initial solutions in map}}
    \State $\mathcal{P} \gets objective\_function(\mathcal{X}_{initial})$ 


      \For{iter = $1 \to I$}
      \Comment{\textit{Create new solutions from elites}}
        \State $\mathbf{x~} \gets random\_selection(\mathcal{X})$
        \State $\mathbf{x'} \gets random\_variation(\mathbf{x})$
        \State $\mathbf{b'} \gets feature\_descriptor(\mathbf{x'})$
        \State $\mathbf{p'} \gets objective\_function(\mathbf{x'})$
        \If{$\mathcal{P}(\mathbf{b'}) = \emptyset$ or $\mathcal{P}(\mathbf{b'}) < \mathbf{p'}$}
        \Comment{\textit{Replace genome if a better one is found}}
          \State $\mathcal{P}(\mathbf{b'}) \gets \mathbf{p'}$
          \State $\mathcal{X}(\mathbf{b'}) \gets \mathbf{x'}$
        \EndIf
      \EndFor
      \State \Return $(\mathcal{X}$, $\mathcal{P})$ 
      \Comment{\textit{Return illuminated map}}
      \EndFunction
  \end{algorithmic}
\end{algorithm}

MAP-Elites has been shown to be effective in finding high-quality diverse solutions in a variety of domains including the design of walking soft robot morphologies~\citep{Mouret2015}, the generation of images that fool deep neural networks~\citep{nguyen2015deep}, and the evolution of robot controllers for damage adaptation~\citep{cully2015robots,chatzilygeroudis2017, pautrat2017}.

SAIL uses MAP-Elites rather than NSLC for illumination. 
Whereas the niche definitions of NSLC are emergent, and so inconsistent across runs, MAP-Elites defines a fixed structure of feature space boundaries, simplifying the process of sampling new solutions for inclusion in the surrogate model. 
Additionally, for design space exploration, this consistency allows designers to easily visualize and compare the effects of altered constraints and conditions on the performance potential of a fixed feature space.

 	\subsection*{Surrogate-Assisted and Bayesian Optimization} \label{sec:bo}

Evolutionary approaches typically require a large number of evaluations before acceptable solutions are found. In many applications these performance calculations are far from trivial, and the overall computational cost of repeated evaluations becomes prohibitively expensive for evolutionary optimization. In these cases approximate models of the fitness function, also known as metamodels or surrogate models, can be used in their place \citep{Emmerich2002, Jin2005}. 
Surrogate-assisted optimization techniques have been particularly important in domains which require complex fluid dynamics simulations to measure performance, in particular aerodynamic design~\citep{Hasenjager2005, Giannakoglou2006,Zhou2007, Dumas2008, Forrester2009, Lian2010, gaier2017aerodynamic}.

Modern surrogate-assisted optimization often takes place within the framework of Bayesian optimization (BO)~\citep{Brochu2010, Calandra2013, cully2015robots, Shahriari2016, pautrat2017}. BO approaches an optimization task not as one of finding the most optimal solution, but of modeling the underlying objective function in high-performing regions. 

Bayesian optimization has two components. The first is a surrogate model of the objective function. 
These surrogates are probabilistic data-driven models based on a set of input/output pairs, or \textit{observation set}. The initial set of solutions can be taken randomly or by sampling the parameter space with design of experiments techniques such as Latin hypercube sampling or Sobol sequences. Sobol sequences~\citep{Niederreiter1988} iteratively sample a multidimensional range such that the range is divided into finer and finer uniform partitions, approximating a uniform sampling. 

The second component of BO is the \textit{acquisition function}, which describes the utility of evaluating the objective function at a given point. BO proceeds by searching for the point with maximal utility, evaluating it on the objective function, and adding this input/output pair to the set of observations. The updated observation set is then used to produce a more informed model. The process then repeats, refining the model of the objective function with each new evaluation.

As the active learning process of BO is dependent on selecting points with the highest utility, how ``utility'' is defined can be critical to the algorithm's performance. Balance must be maintained between exploration, evaluating points with high uncertainty, and exploitation, evaluating points which are likely to have high fitness.
Choosing new points to evaluate based solely on predicted fitness is too greedy, and will result in premature convergence on local optima. At the other extreme, evaluating points with the least confidence will decrease the uncertainty of our models globally, but it is also wasteful: with only limited resources improving accuracy around optima should be prioritized, not the search space as a whole.
New points for evaluation should be chosen where the model predicts both high fitness and where model uncertainty is high. How this balance is struck is defined precisely by the acquisition function.

The \textit{upper confidence bound} (UCB)~\citep{Srinivas2009} is an intuitive and straightforward acquisition function. UCB judges new points optimistically, favoring uncertainty under the assumption that higher uncertainty hides a potentially higher reward. UCB can be defined as a weighted sum of the mean~($\mu$) and uncertainty~($\sigma$) of the prediction, where both a high mean and large uncertainty are favored, and their relative emphasis tuned by the parameter $\kappa$:
	\begin{equation}
	  UCB(\mathbf{x}) = \mu(\mathbf{x}) + \kappa\sigma(\mathbf{x})
	\end{equation}

Proposed as part of the GP-UCB algorithm, use of UCB has been shown to minimize regret and maximize information gain in multi-armed bandit problems~\citep{Srinivas2009}, and performs competitively with more complex acquisition functions such as Expected Improvement (EI) and Probability of Improvement (PI) \citep{Brochu2010,Calandra2013}.

 	\subsubsection*{Gaussian Process Models}\label{sec:gp}
Surrogate models can be constructed using any number of data-driven machine learning techniques, including polynomial regression, support vector machines, and artificial neural networks~\citep{Jin2005,Forrester2009}. For BO, because a probabilistic prediction is required to assess uncertainty, Gaussian process~(GP) models~\citep{Rasmussen2006} are typically used.

GP models are accurate even with small data sets and their predictions include a quantified level of certainty. In the active learning context of surrogate-assisted optimization a measure of model uncertainty is particularly useful, as this allows for the balancing of exploration and exploitation.

Gaussian process models use a generalization of the Gaussian distribution: where a Gaussian distribution describes a distribution of random variables, defined by mean and variance, a Gaussian process describes a random distribution of functions, defined by a mean function $\mu$, and covariance function $k$. 

\begin{equation}
f(x) \sim GP(\mu(x), k(x,x'))
\end{equation}

In much the same way as an artificial neural network can be thought of as a function that returns a scalar given an arbitrary input vector $x$, a GP model can be thought of as a function that, given $x$ returns the mean and variance of a normal distribution, with the variance indicating the certainty of the prediction.

Gaussian process models make their predictions based on locality in the input space, a relationship defined by a covariance function. A common choice of kernel for this covariance function is the squared exponential function: as points become closer in input space they become exponentially correlated in output space.

\begin{equation} \label{eq:sqExp}
k(\mathbf{x_i,x_j}) = \exp{ \Big( -\frac{1}{2} \|\mathbf{x}_i - \mathbf{x}_j\|^2 \Big)}
\end{equation}
Given a set of observations $D = ({x_{1:t}, f_{1:t}})$ where $f_{1:t} = f(x_{1:t})$,  we can build a matrix of covariances. In the simple noise-free case we can then construct the kernel matrix:
\begin{equation}
K = 
 \begin{bmatrix}
  k(x_1,x_1) 	& \cdots  	& k(x_1,x_t) 	\\
  \vdots  		& \ddots  	& \vdots 		\\
  k(x_t,x_1) 	& \cdots  	&  k(x_t,x_t)
 \end{bmatrix}
\end{equation}

Considering a new point ($x_{t+1}$) we can derive the value ($f_{t+1} = f(x_{t+1})$) from the normal distribution:
\begin{equation}
P(f_{t+1}|D_{1:t},x_{t+1}) = 
\mathcal{N} \Big( \mu_t (x_{t+1}), \sigma_t^2 (x_{t+1}) \Big)
\end{equation}
where: 
\begin{align}
\mu_t(x_{t+1}) &= \mathbf{k}^T \mathbf{K}^{-1} \mathbf{f}_{1:t} \\
\sigma_t^2(x_{t+1}) &= k(\mathbf{x}_{t+1}, \mathbf{x}_{t+1}) - \mathbf{k}^T \mathbf{K}^{-1}\mathbf{k}
\end{align}
gives us the predicted mean and variance for a normal distribution at the new point $x_{t+1}$. If we then evaluated the objective function at this point, we would add it to our set of observations $D$, reducing the variance at $x_{t+1}$ and at other points near to $x_{t+1}$.

The kernel described in Equation \ref{eq:sqExp} applies a squared exponential relationship to the total euclidean distance between points. In practice, any available domain knowledge should be integrated into the kernel. The kernel need not represent an isotropic distance measure either, and in higher dimensional problems the relative weight of each dimension can be trained along with other hyperparameters when maximizing the likelihood of the model, a technique known as automatic relevance detection (ARD)~\citep{Rasmussen2006}. ARD kernels not only result in a more accurate model, but also supply a human readable estimate of the correlation of each dimension with performance. In our experiments a squared exponential kernel with ARD was used, but any other kernel could be used in SAIL without modification.

\section{Surrogate-Assisted Illumination}

We introduce Surrogate-Assisted Illumination (SAIL) as a Bayesian optimization corollary for quality-diversity. While the goal of BO is to model the behavior of the objective function in the highest performing region, illumination expands this requirement. The goal of illumination is not to find a single optimum, but optima with every combination of features: not a single point, but a slice with one dimension per feature. The goal of SAIL is then to predict the behavior of the objective function in the high-performing regions of this feature slice. Producing models which can perform these predictions requires finding high-utility solutions which cover these feature dimensions.


These high-utility solutions are found by MAP-Elites. Maximizing the acquisition function in every feature bin produces an \textit{acquisition map}, a set of high-utility solutions that span the feature dimensions. As utility is derived using only the Gaussian process model, an acquisition map can be produced with minimal computation. To place solutions in the map requires that features as well as fitness be calculated. In design cases, features can typically be cheaply derived without evaluation. In some domains, such as evolutionary robotics, it may not be possible to cheaply extract features, as they are often behaviors exhibited during evaluation. In such cases, features as well as fitness would need to be approximated.

To define utility SAIL uses the UCB acquisition function (see Section~\ref{sec:bo}: \textit{Surrogate-Assisted and Bayesian Optimization}) rather than other common acquisition functions such as Expected Improvement (EI) and Probability of Improvement (PI)~\citep{Brochu2010,Calandra2013}. These acquisition functions rely on comparisons to the current optimum, while UCB is based only on the confidence of the underlying model. As SAIL solves numerous localized problems in parallel, it requires an acquisition function independent of the global optimum. If compared globally, solutions in less optimal regions of the map would have a vanishingly small probability of improving on the global optimum, and because many bins will not contain existing evaluated solutions, it will not always be possible to perform local comparisons against optima within a bin.

The acquisition map acts as a collection of candidate solutions for evaluation. As our goal is to create a model that is accurate on the entire feature slice, it is necessary to evaluate high-utility solutions which cover the slice in its entirety, not only at the highest utility point. Just as a Sobol sequence can be used to evenly sample the parameter space, it can also be used to evenly sample the acquisition map. Feature coordinates are drawn from a Sobol sequence, and the elite contained in the corresponding bin of the acquisition map is selected for evaluation. Any number of new solutions from the acquisition map can be selected by drawing the next sets of coordinates from the Sobol sequence. The selected solutions are then evaluated on the objective function, and the resulting input/output pairs added to the model. The illumination process is then repeated with this more accurate model, creating a new acquisition map.

The model refined through this iterative illumination and modeling process can be used to produce a \emph{prediction map}. By adjusting the acquisition function so that it does not reward uncertainty, MAP-Elites will produce a map which includes the model's best guess of the optimal design in each bin. This map provides an informed estimate of the relationship between features and performance, and as only the surrogate model is required, this prediction map can be produced with minimal computation.


\begin{algorithm} [h]
  \caption{Surrogate-Assisted Illumination (SAIL)}
  \label{alg:sail}

  \begin{algorithmic}[1]
    \State{\textbf{1) Create Gaussian Process Model}}
    \State $\mathcal{X} \gets Sobol_{1:G}$ \Comment{\textit{Initialize with G solutions drawn from Sobol sequence}}
    \State $\mathcal{P} \gets PE(\mathcal{X} )$ \Comment{\textit{Precisely Evaluate (PE) solutions to get performance}}
        \State $\mathcal{GP} \gets Gaussian\_process\_model(\mathcal{X}, \mathcal{P})$ \Comment{\textit{Train GP model}}

    \\\State{\textbf{2) Produce Acquisition Map}}
    \While{\textit{precise evaluation budget not exhausted}}
        \State $acquisition() \gets UCB(\mathcal{GP}(x))$ \Comment{\textit{Use UCB of prediction as fitness function}}
        \State $(\mathcal{X}_{acq}, \mathcal{P}_{acq}) =$ 
        \Call{MAP-Elites}{$acquisition(), \mathcal{X}$}
        \Comment{\textit{Create acquisition map}}
        \State $\mathbf{x} \gets \mathcal{X}_{acq}(Sobol_{iter})$ 
        \Comment{\textit{Select solutions from acquisition map for PE}}
        \State $\mathcal{X} \gets \mathcal{X} \cup \mathbf{x}$,
        $\mathcal{P} \gets \mathcal{P} \cup PE(\mathbf{x})$
        \Comment{\textit{Add evaluated solutions to observation set}}
            \State $\mathcal{GP} \gets Gaussian\_process\_model(\mathcal{X}, \mathcal{P})$ \Comment{\textit{Train GP model}}
    \EndWhile

    \\\State{\textbf{3) Produce Prediction Map}}
    \State $prediction() \gets mean(\mathcal{GP}(x))$ \Comment{\textit{Use mean of prediction as fitness function}}
    \State $(\mathcal{X}_{pred}, \mathcal{P}_{pred}) =$ 
    \Call{MAP-Elites}{$prediction(), \mathcal{X}$}
    \Comment{\textit{Create prediction map}}

  \end{algorithmic}
\end{algorithm}

The SAIL algorithm is more precisely defined in Algorithm~\ref{alg:sail}. An initial set of individuals is created using a Sobol sequence~\citep{Niederreiter1988} to ensure our model is based on solutions which evenly cover the \emph{parameter} space. These individuals and their performance form the set of observations which are used to construct the initial GP model. An empty acquisition map is then created and filled with the individuals from the observation set, along with their utility as judged by the acquisition function. These individuals are taken as the starting population for MAP-Elites (see Algorithm~\ref{alg:mapelites}) which then illuminates the map as described in Section~\ref{sec:qd}: an elite is selected and mutated to produce a child, it is assigned a bin based on its features, and then competes for the bin if it is not occupied. This illumination process repeats for a number of iterations, and results in an acquisition map of elite individuals who maximize the acquisition function in their bin. A \textit{prediction map} can then be produced by maximizing only the predicted mean performance in each bin with MAP-Elites, using the evaluated individuals as a starting population.

 \label{sec:sail}

 \section{Summary of 2D Airfoil Results} \label{sec:2d}




In \cite{gaier2017data} the capabilities of SAIL were demonstrated on the classic design problem of airfoil shape optimization. SAIL is designed with computationally intensive domains in mind, but this relatively inexpensive domain allows for deeper analysis and evaluation of the algorithm's performance. The low computational cost of evaluations allowed us to demonstrate that: 
(1) the designs found by SAIL are near optimal in all regions of the feature space, 
(2) the models created by sampling with SAIL are an order of magnitude more accurate in high-fitness regions than those produced by uniform sampling of the parameter space, and that 
(3) SAIL produces high-performing solutions in every feature region with the same computational budget required by a standard black-box optimizer to find a single design. 
As the computational cost of simulation in the three-dimensional aerodynamics case presented in the next section prevents this level of analysis, we summarize the results of \cite{gaier2017data} here.

\subsection{2D Airfoil Optimization: Objectives and Comparisons} \label{sec:2dsetup}

SAIL is given the task of producing a set of 2D airfoils with minimal drag which still maintain the lift and area of the high-performing RAE2822 airfoil. As both the drag and lift of a given airfoil must be approximated, and no correlation between the two are assumed, two GP models are used to produce a fitness estimate. In both models a squared exponential kernel with ARD (see Section \ref{sec:gp}, \textit{Gaussian Process Models}) is used.
The first model, used to estimate drag, performs regression and rewards both high mean\footnote{drag was measured as $ - \log ( C_D (x)) $, i.e. higher values correspond to lower coefficients of drag} and high variance values in the UCB acquisition function. 
The second, used to estimate lift, performs classification using the mean and variance given by the GP model to return a probability that the lift is above the constraint threshold. This probability is applied as a penalty to the estimated fitness. 
The area of a foil is directly measured, with any deviation from the area of the RAE2822 airfoil penalized in the fitness function. The resulting fitness of a design $x$ is then:

\begin{equation}
\mathit{fitness}(x) = (\mu_{\text{drag}}(x) + \kappa\sigma_{\text{drag}}(x) ) \times \mathit{penalty}_{\mathit{lift}}(x) \times \mathit{penalty}_{\mathit{area}}(x)
\end{equation}
where 
$\mu_{\text{drag}}(x)$ is the prediction mean,
$\sigma_{\text{drag}}(x)$ is the prediction uncertainty, and
$\kappa$ is the UCB weighting coefficient. For more detailed definitions of setup and fitness derivation see \cite{gaier2017data}.

\begin{figure}[ht]
	\centering
	\includegraphics{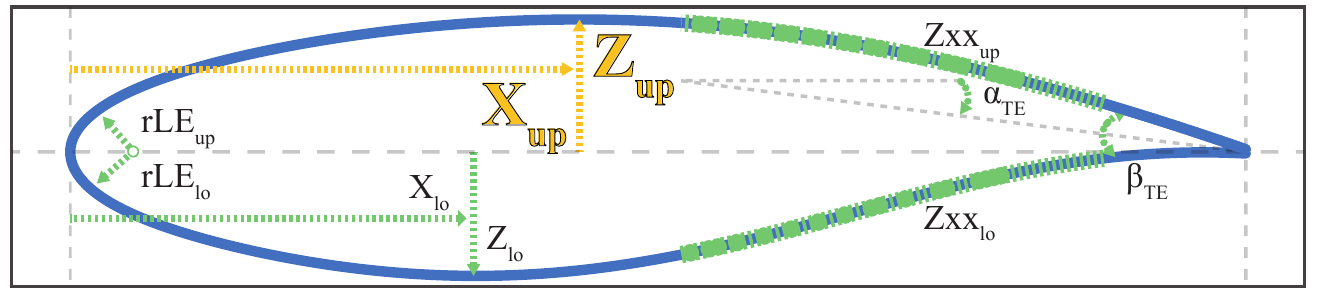}
	\caption
	{\textit{The Ten Parameters Used to Define an Airfoil} \newline
		Airfoils are defined by a set of features using the PARSEC encoding. These features include: the leading edge radius of the upper and lower curves ($rLE_{up}$, $rLE_{lo}$), the location of highest and lowest point ($Z_{up}$, $Z_{lo}$), the location of highest and lowest point ($X_{up}$, $X_{lo}$), the curvature of the upper and lower sides ($Zxx_{up}$, $Zxx_{lo}$), and the angle and arc radius of the trailing edge ($\alpha_{TE}$, $\beta_{TE}$). The dimensions of variation explored ($X_{up}$ and $Z_{up}$) are colored in gold.}
	\label{fig:parsec}
\end{figure}

Airfoils are encoded using the airfoil-specific PARSEC parameterization~\citep{Sobieczky1999}. PARSEC allows the direct parameterization of features such as the radius of the leading edge or the curvature of the upper surface, and so allows a large variety of designs to be expressed with a small number of parameters. The ten parameters used to define an airfoil in these experiments are shown in Figure~\ref{fig:parsec}. 

In these experiments two PARSEC descriptors, the height of the airfoil ($Z{up}$) and the location along the airfoil of that highest point ($X{up}$) are used as features of variation for exploring the design space. Selecting parameter values as features of variation allows the results of SAIL to be compared with those of standard optimization algorithms designed to find a single solution. These algorithms have no concept of `features', but by restricting the search within given parameters ranges, the regions of feature space explored can also be restricted.

SAIL is compared to (1) the original MAP-Elites algorithm, (2) the black-box optimization algorithm Covariance Matrix Adaptation Evolution Strategy (CMA-ES)~\citep{Hansen2001}, and (3) a surrogate-assisted variant of CMA-ES (SA-CMA-ES).
As the primary objective of SAIL is to achieve results with data-efficiency, the unit of comparison used is the number of function calls, or precise evaluations (PE) required to achieve a certain result. SAIL is given a total computational budget of 1000PE. 50PE is used to evaluate a set of designs produced through parameter sampling, which forms the basis of the initial GP model. At each acquisition iteration (Algorithm \ref{alg:sail}: lines 6-13) 10 additional individuals are chosen from the acquisition map, evaluated, and incorporated into the GP model until the evaluation budget is exhausted. This result is compared with the designs produced by the MAP-Elites algorithm when it is given a computational budget of $10^5$PE.

 Optimality of designs is judged in comparison to the black-box optimization algorithm CMA-ES. By restricting the allowed parameter ranges, the search is confined to a single feature bin, and CMA-ES is given a budget of 1000PE to produce a design in each bin. To control for the ease of modeling the problem we also employ a surrogate- assisted variant of CMA-ES, denoted SA-CMA-ES, to solve each of these subproblems. Within a single bin 25 initial solutions are sampled from the parameter space and used to produce an initial GP model. CMA-ES is then used to maximize the same UCB-based acquisition function used by SAIL, described above. The single found optimum is evaluated and the new individual incorporated into the GP model. This process is repeated for a total of 100PE. 

The 2-dimensions of the feature space ($Z_{up}$ and $X_{up}$) are each divided into 25 partitions, for a total of 625 bins. Bins with the highest point at the trailing edge of the wing (high $Z_{up}$ and low $X_{up}$) could not be found due to geometric constraints inherent to the PARSEC representation~\citep{Padulo2009}. Only the remaining $577$ bins were considered in statistical comparisons. Each approach was replicated 20 times with different random seeds.\footnote{One replicate, including precise evaluation of all designs in the intermediate prediction maps used for data gathering, with 8 cores of a Intel Xeon 2.6GHz processor required:
SA-CMA-ES:32h, CMA-ES:80h, SAIL:12h, MAP-Elites:14h Where possible, standard implementations of algorithms were used, including the CMA-ES as published by Hansen~\citep{cmaes}, the Gaussian processes for machine learning toolbox by Rasmussen~\citep{gpml}, and the XFoil airfoil solver of Drela~\citep{xfoil}.}
Unless otherwise mentioned, all values given are medians over all replicates. 

\newpage
\subsection{2D Airfoil Optimization: Exploration}

\begin{figure}[ht]
	\centering
	\includegraphics{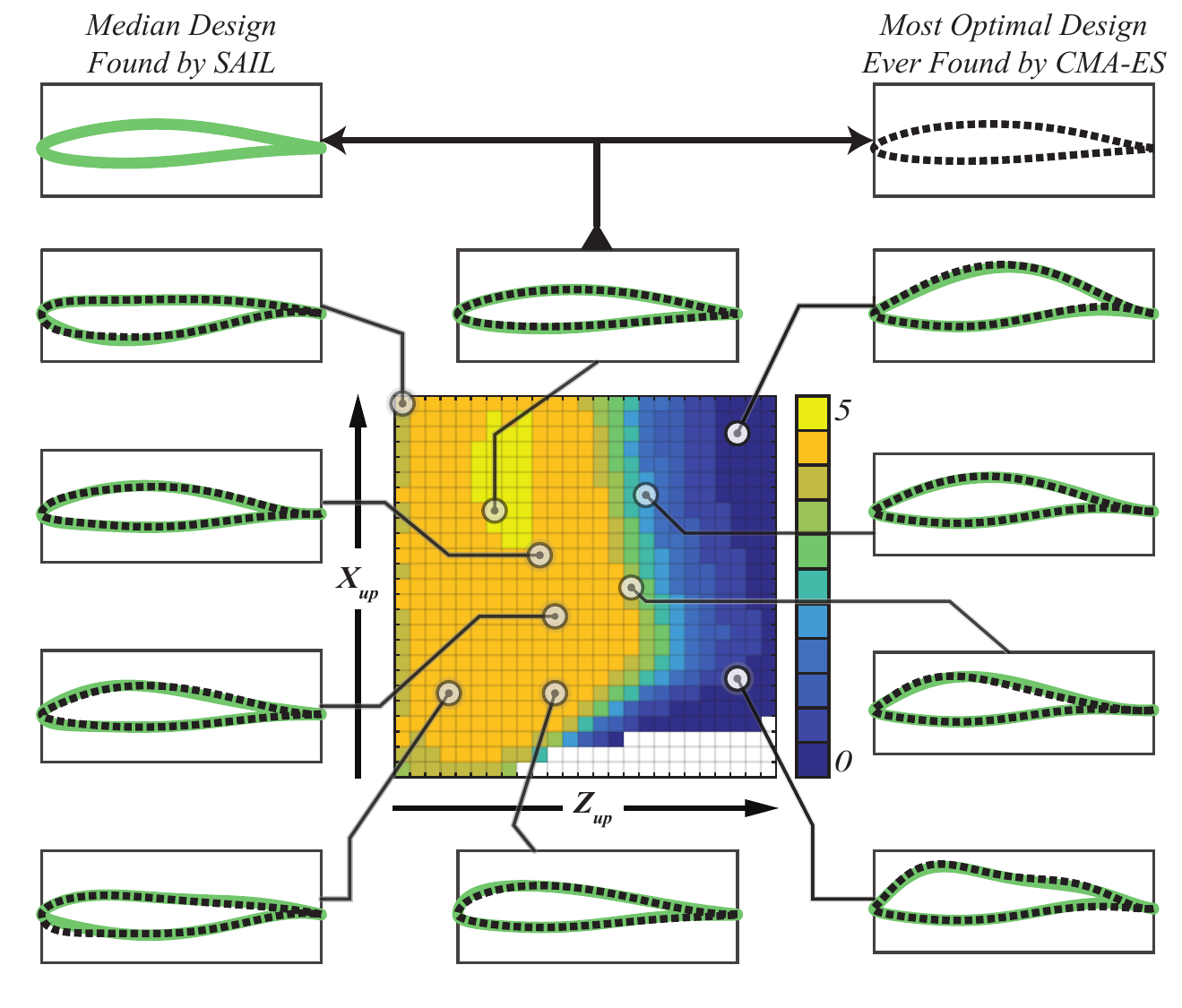}
	\caption{
		\textit{Design Space Overview Produced with Sail}
		\newline Prediction map produced by SAIL after 1000PE, bins colored by fitness. Fitness is dominated by the drag component: $-log(C_D)$, e.g. a fitness of 5 is equivalent to a $C_D$ of $10^{{-5}}$. The border contains designs in selected bins: median-performing designs found by SAIL in green, best designs ever found in \emph{all} CMA-ES runs in black. 
		}
	\label{fig:designMap}
\end{figure}

In Figure~\ref{fig:designMap} the median-performing prediction map produced by SAIL is shown. When each bin is color coded by the fitness of the design within, it provides an intuitive overview of the relationship between airfoil features and performance: the height of the airfoil ($Z{up}$) has the most influence, with the location of the highest point ($X{up}$) having a more nuanced effect, largely dependent on the height of the airfoil. The solutions in this map represent the model's best guess of the optimal designs across the feature space. When each design is evaluated in the simulator, we find that 90\% of our model's drag predictions for these optimal designs are within 5\% of their true value \citep{gaier2017data}. 
Around the border the median-performing design found by SAIL in a bin is shown in green, along with the most optimal design ever found by CMA-ES in black. The designs are very similar, though found by SAIL in a single 1000 evaluation run and by CMA-ES over many runs using more than 11.5 million evaluations\footnote{20 replicates $\times$ 577 bins $\times$ 1000 precise evaluations}. 

\subsection{2D Airfoil Optimization: Accuracy}

The models produced by SAIL are designed to accurately approximate solutions in a high-fitness slice across the feature space. Comparing the predicted and true performance of designs in the prediction map can tell us how accurate our predictions are, but not necessarily how accurately we model the high-fitness slice. To measure how well our models predict performance in high-performing regions they must be tested on the true optimal solutions, not on the solutions produced by SAIL.

By considering the best-performing designs found by CMA-ES over all replicates as representative of this high-fitness slice, we can examine how the accuracy of our model's predictions improve as they gather more samples. In SAIL we guide selection toward promising and unexplored regions by searching for solutions which maximize the upper confidence bound (UCB), a weighted sum of the mean and variance of the prediction. To illustrate the value of this approach, we examine the effect when SAIL instead uses an acquisition function of \emph{only} the mean or \emph{only} the variance. As a baseline we compare the models built using sampling guided by SAIL to those built through even parameter sampling with a Sobol sequence~\citep{Niederreiter1988}. As in this particular airfoil case the features are themselves parameters, this Sobol sampling will, as SAIL, provide even sampling across the feature space.

\begin{figure}[h!]
\centering
	\includegraphics{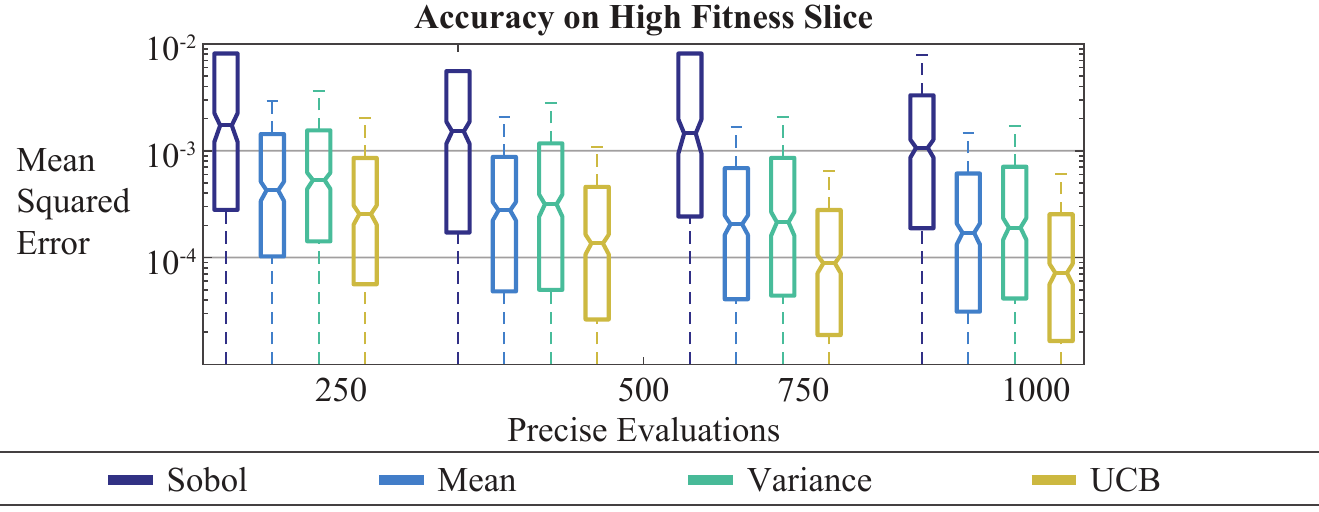}
	\caption
	{
		\textit{UCB Sampling Outperforms Sampling for Performance or Variance Alone}\newline
		Mean squared error (log scale) of drag prediction on optimal designs found by CMA-ES. Models were constructed using designs: 1) sampled from parameter space using a Sobol sequence, or selected from acquisition maps where individuals were optimized to maximize their predicted 2) mean, 3) variance, or 4) Upper Confidence Bound (UCB). Though all acquisition functions produced more accurate models than uniform parameter sampling, UCB improved by a full order of magnitude on high-fitness solutions.
	}
	\label{fig:modelComparison} 
\end{figure}

The accuracy of each resulting model's drag predictions on the high-fitness slice at various stages of SAIL's acquisition process is shown in Figure~\ref{fig:modelComparison}. We see that using the SAIL algorithm to select samples from across the feature space improves accuracy over evenly sampling the parameter space, regardless of whether uncertain or high-performing solutions are favored. When both the uncertainty and performance are considered with UCB, SAIL produces models which are an order of magnitude more accurate on the high-fitness slice than uniform sampling of the parameter space.

\subsection{2D Airfoil Optimization: Performance}

SAIL is designed as a data-efficient variant of MAP-Elites, and so its ability to find optimal solutions must be examined in this context. Beyond MAP-Elites there are few feature space exploration algorithms, and so for additional comparison we look to the standard black-box optimizer CMA-ES. As CMA-ES is not designed for use across a multitude of subproblems, the number of evaluations required to produce an optimized feature map is highly dependent on the number of bins in the map.

More informative than comparing performance across the entire map is to compare SAIL to the performance of CMA-ES within a single bin. While it is not expected that SAIL will compete with an algorithm like CMA-ES in finding a single solution, it allows us to put SAIL's optimization performance in context. As optimization progress may vary according to the bin, we take the single bin performance as the map performance divided by the number of bins. We include a surrogate-assisted variant of CMA-ES (SA-CMA-ES) ({see~Section~\ref{sec:2dsetup}}) to control for the ease of modeling the problem.

\begin{figure}[h]
	\centering
	\includegraphics{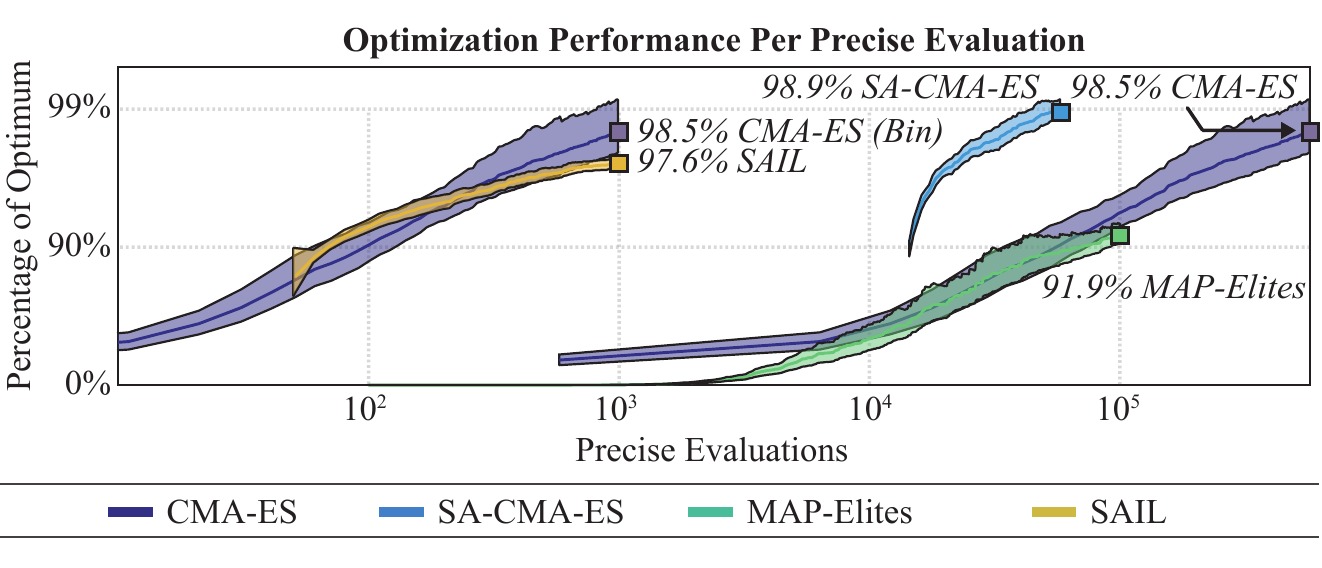}
	\caption[caption]
	{
		\textit{Data-efficiency of Optimization Over the Entire Design Space}\newline
		Computational efficiency (log scale) of CMA-ES, SA-CMA-ES, SAIL, and MAP-Elites measured in precise evaluations. Median performance across all bins as percentage of optimum (log scale). For SAIL and SA-CMA-ES, performance is only recorded after initial sampling. As the optimum is the highest performance over \emph{all} runs, the median CMA-ES run does not reach this value.
		\emph{Bin}: median progress towards optimum of every bin. Bounds indicate one standard deviation over 20 replicates.
	}
	\label{fig:progress}
\end{figure}

Figure~\ref{fig:progress} illustrates the comparison of performance per precise evaluation. With the same evaluation budget required by CMA-ES to find a near optimal solution in a \textit{single} bin, SAIL finds designs of comparable performance in \emph{every} bin. Comparisons between CMA-ES and MAP-Elites and their surrogate-assisted variants SA-CMA-ES and SAIL reveal that the gains from surrogate-assisted optimization are even greater for MAP-Elites than for the traditional optimizer. While surrogate-assistance improves the efficiency of CMA-ES by an order of magnitude, even when MAP-Elites is given an evaluation budget two orders of magnitude greater than SAIL it cannot produce solutions of similar performance.

\section{Illumination of 3D Aerodynamics Design Spaces}


	To further explore the capabilities of SAIL, we choose a more demanding task: the optimization of aerodynamic shells for recumbent bicycles, or velomobiles. These streamlined vehicles hold human-powered speed (144.17 km/h, \cite{ihpva}) and distance (680 km in 12 hours, \cite{whpva}) records due to their highly tuned aerodynamics.

	\begin{figure}[h!]
	\centering
	\includegraphics[width=.8\textwidth]{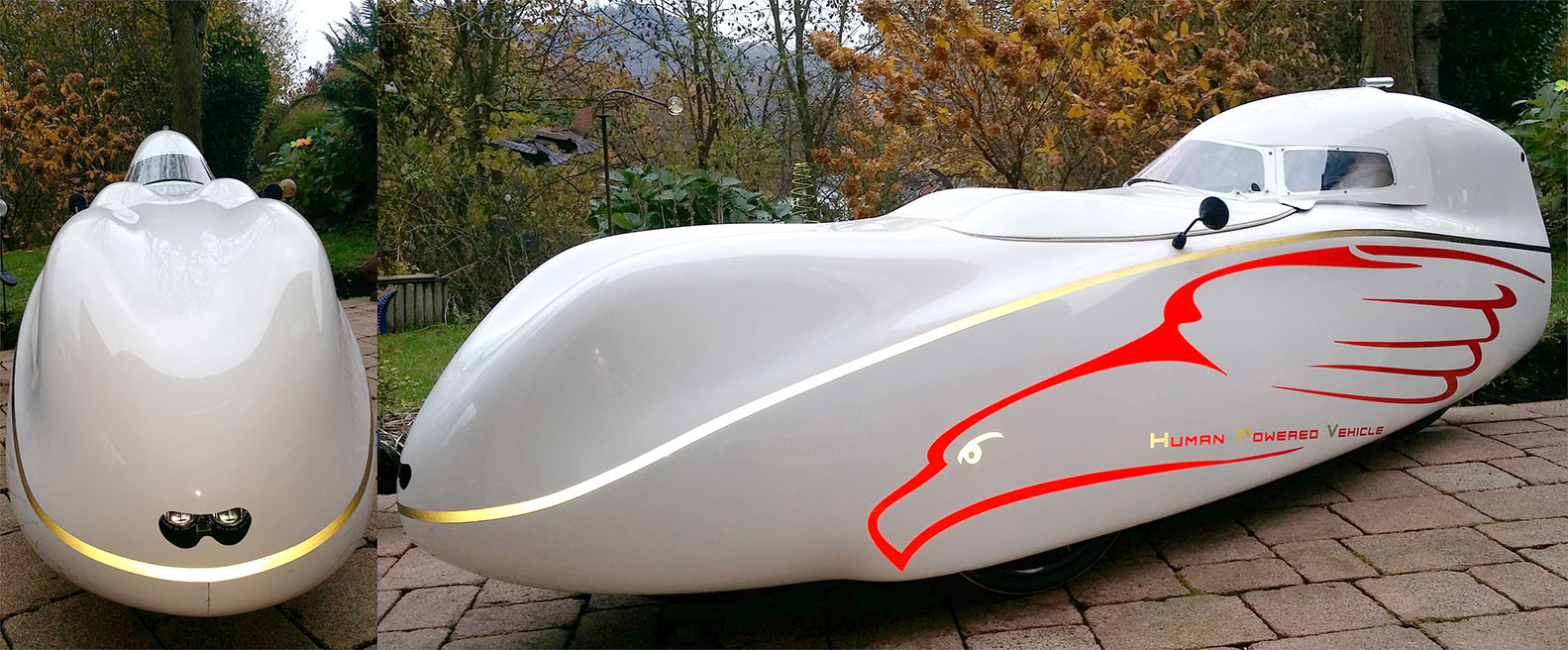}
  	\caption{\textit{Record Setting Velomobiles Have Non-intuitive Shapes.}\newline The record setting Milan velomobile eschews the intuitive bullet shape in favor of a form which decreases frontal area while still providing space for the riders feet and knees. Bumps on the sides guide and direct the flow that splits over this uneven surface, reduces the effects of side winds, and adds rigidity to the thin carbon fiber shell.}
	\label{fig:milan}
	\end{figure}

	Due to constraints such as rider movement and comfort, three-wheeled designs built for distance races often have non-intuitive shapes (Figure \ref{fig:milan}). These high-performing but odd designs suggest a domain rich in interesting design concepts. To encode the design of the shell a parameterized encoding could be created, as was done with PARSEC for airfoils, or a more general-purpose solution could be applied, by deforming an existing design. To isolate the capabilities of SAIL from the capabilities of a given representation we examine and compare the result for both encoding approaches.


	While the PARSEC representation is composed of specifically engineered features that are known to be important for the performance of airfoils, most design domains have not been as intensely researched. It is more useful to be able to define a realistic set of features which do not directly correspond to the parameter values of the representation. Here we explore the curvature and volume of the designs, neither of which directly correspond to parameter values of our encodings.
	
	Illuminating the design space according to features which are unaligned with the parameters of the encodings demonstrates that SAIL is doing more parallel search in a partitioned search space. SAIL produces designs that vary across a spectrum in a low-dimensional feature space, illuminating the relationship between features and performance in a way largely independent of the method by which the designs are encoded.



	\subsection{Encodings}
		\subsubsection*{Parameterized Design} 

To produced a parameterized encoding for the smooth form of the velomobile shell, we use a series of 2D airfoil-like curves defined with the PARSEC encoding~\citep{Sobieczky1999} (\emph{see Section \ref{sec:2dsetup}: 2D Airfoil Optimization}).

\begin{figure}[ht!]
	\centering
	\includegraphics{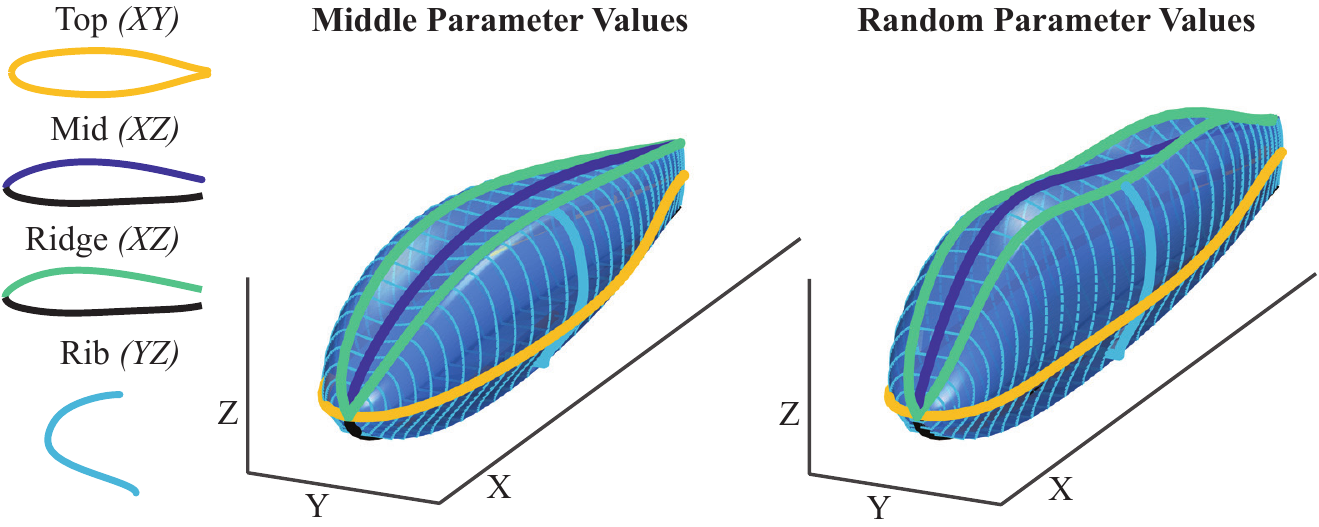}
  	\caption{
  	\textit{Velomobile Design in 3D Produced from a Set of 2D PARSEC Curves} \newline
  	PARSEC curves are defined in two dimensions (\textit{left}) and connected by splines. Top (XY) profile in yellow, side profiles (XZ) for the middle profile in purple, with raised knee ridges in green, the front (YZ) profile rib in cyan. This rib connects the side profile to a fixed base, in black, and is scaled to fit the width of the top profile at every section. The resulting 3D designs are shown when parameter values are set to the middle of the range (\textit{middle}) and when random parameters are chosen (\textit{right}).}
	\label{fig:paramRep}
\end{figure}

\begin{table}[ht!]
	\centering	
	\begin{tabular}{r||l|l|l|l}
	\multicolumn{1}{c||}{{ Curve}} & \multicolumn{4}{c}{{ PARSEC Parameter}}                                      \\ \hline
	\rowcolor[HTML]{EFEFEF} 
	\textit{Top (XY)}                      & $r_{\text{LE}}$    & $X_{up}$ & $Z_{up}$   & $Zxx_{up}$ \\
	\textit{Mid (XZ)}                      & $r_{\text{LE}}$    & $X_{up}$ & $Z_{up}$   & $Zxx_{up}$ \\
	\rowcolor[HTML]{EFEFEF} 
	\textit{Ridge (XZ)}                    & $r_{\text{LE}}$    & $X_{up}$ & $Z_{up}$* &           \\
	                                  & $r_{\text{LE}}$    & $X_{up}$ & $Z_{up}$  & $Zxx_{up}$ \\
	\multirow{-2}{*}{\textit{Rib (YZ)}}    & $\alpha_{TE}$ &                           &                 &           \\ 
	\end{tabular}
	\caption{
	\textit{The 16 Parameters Used to Determine the Shape of a Velomobile Shell.} 
	\newline Each PARSEC airfoil can be described by 10 or more parameters, but as we are only producing one curve rather than an entire foil, only the parameters which describe the top side are necessary. In addition, we adjust only a limited set of parameters for a total of 16 degrees of freedom. \newline(*) The height of the ridge is defined in relation to the height of the middle profile. }
	\label{table:veloParams}
\end{table}

Viewed from the top (\textit{XY}), the vehicle is defined as a symmetrical airfoil (Figure~\ref{fig:paramRep}, \textit{Top}). The side profile (\textit{XZ}) is defined by three curves: one curve, unchanged in every design, which defines the bottom of the vehicle, and two parameterized curves for the top, one in the center of the vehicle (Figure~\ref{fig:paramRep}, \textit{Mid}) and one which forms a \textit{Ridge} for the knees (Figure~\ref{fig:paramRep}, \textit{Ridge}). A final curve connects this ridge to a flat bottom forming the view from the front (\textit{YZ}) (Figure~\ref{fig:paramRep}, \textit{Rib}). This \textit{Rib}  is defined along a unit vector, and is scaled to remain consistent with the curves defined by the \textit{Top} and \textit{Ridge} curves along 32 sections. Ridges follow the same curve as the \textit{Top} curve of the body. These 2D curves and how they are composed into a 3D shape is illustrated in Figure~\ref{fig:paramRep}. All curves are connected to each other by splines. We only permit a subset of the curve parameters to be modified, limiting the number of representation parameters to 16. These degrees of freedom are enumerated in Table~\ref{table:veloParams}.

		\subsubsection*{Free Form Deformation}

	As an alternative to the hand-designed parameterized encoding, we employ a deformation approach to design optimization that uses free-form deformation (FFD)~\citep{sederberg1986free}. FFD is a well-established technique in computer graphics and design, including evolutionary aerodynamic design optimization~\citep{Samareh1999, Menzel2005, Sieger2012}, which, unlike parameterized representations, allows the designer to begin with a prior design, such as an existing high-performing design, and further refine it. Deformations decouple the complexity of the design from the complexity of the representation: an intricate hand designed object can be deformed and optimized with only a few degrees of freedom, without the need to design a representation which can recreate the original design. In fluid dynamics applications meshes must be created for every design and this meshing is itself a difficult and time consuming process, which in many cases must be done by hand. However, when deformations are applied the mesh does not have to be recreated but instead likewise deformed, greatly simplifying the process, and so this onerous step can often be omitted.	

	To perform FFD on a design, it is first embedded into a lattice of control points. The mesh points of the design are converted into a local coordinate system based on these control points, so that any point $X$ has $(s,t,u)$ coordinates in lattice space 

	\begin{equation}
	X(s,t,u) = X_0 + sS + tT + uU
	\end{equation}

	where $X_0$ is the origin in lattice space and $S,T,U$ dimensions that lie along the edges of the control volume. For any point inside the lattice $s$, $t$, and $u$ are between $0$ and $1$. Control points are defined in a grid along the control volume as:

	\begin{equation}
	P_{\text{ijk}} = X_0 + \frac{i}{l}S + \frac{j}{m}T + \frac{k}{n}U
	\end{equation}

	where $l$, $m$, $n$ are the number of control points in the $S$, $T$, and $U$ dimensions. When a control point is moved, these mesh points are also adjusted to maintain their position in relation to the control points. The influence of each control point on a point in the mesh is determined by a Bernstein polynomial blending function $B$. To get the deformed position in Cartesian space of a given point $X_{\text{ffd}}$, we first convert the point's location to $(s,t,u)$ coordinates then compute shifts based on each control point: 
	\begin{equation}
	X_{\text{ffd}} = \displaystyle\sum_{i=1}^{l} \displaystyle\sum_{j=1}^{m} \displaystyle\sum_{k=1}^{n} P_{\text{ijk}} B(s) B(t) B(u) 
	\end{equation}

	\begin{figure}[h]
	\centering
	\includegraphics{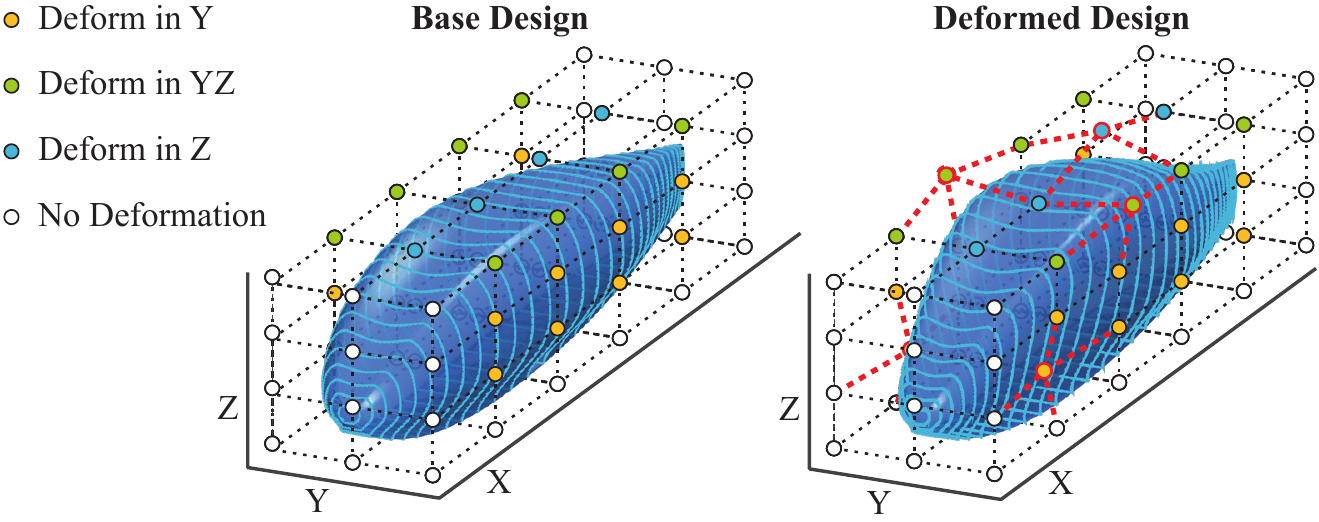}
  	\caption{\textit{Velomobile Design Created through Free-Form Deformation.}\newline
  	A control volume of evenly space points is constructed around the base velomobile shape. Active control points are shown in color: orange points deform along the Y axis, blue points along the Z axis, and green points move equally in the Y and Z axis. All points are stationary in the X dimension. Orange and green points move in tandem with those on the other side of the velomobile to enforce symmetry, resulting in a total of 16 degrees of freedom.\newline
  	\textit{Left:} The base design with no deformation, \textit{Right:} A new design is created by adjusting three parameters which, once symmetry is enforced, moves two control points in Y, two control points in the YZ plane, and one control point in the Z dimension.
  	}
	\label{fig:ffdRep}
	\end{figure}

	We design the control lattice and degrees of freedom of our FFD encoding in such a way as to keep it comparable to our parameterized representation. As a base shape we use the design produced by the parameterized encoding with every parameter at the center of its parameter range. Deformation control points can move in the positive or negative direction, and here are normalized so that at the center of the deformation range no deformation takes place. It follows then that when all parameter values of both encodings are at the center of their range, they produce identical designs. Our control lattice is constructed in an intuitive manner, surrounding the design in its entirety, with each dimension corresponding to one in Cartesian space. 

	Control points are placed to evenly divide the area surrounding the design into 6 segments in the X axis, 3 segments in Y, and 4 in Z (see Figure~\ref{fig:ffdRep}). We only actively manipulate a subset of the control points, and these we only move in a single dimension. The control points at the bottom of the velomobile are left unmoved, restricting deformations of the base, as is done in the parameterized case. The control points at the front and back of the design are left unmoved as well, keeping the start and end points of the design in line with that of the parameterized encoding. We allow the 4 center top control points to move up and down in the Z dimension, the 8 center side control points to move in and out in the Y dimension, and the 4 top corner points to move in both the Y and Z dimension, though only at the same rate. Enforcing symmetry in the Y dimension results in a total of 16 degrees of freedom, as in the parameterized case.

	\subsection{Features}

Two dimensions of variation are explored: volume and curvature. While it is obvious that lower volume designs will produce less drag, it is just as obvious that a design with no volume is not optimal. A designer could determine the specific dimensions for a given configuration of machinery and rider, and then codify these as constraints for an optimization algorithm. 
SAIL instead produces a set of high-performing designs of varying volumes which the designer can browse. A design can then be selected which satisfies their constraints, or which could satisfy them with small adjustments. 

\begin{figure}[h!]
	\centering
	\includegraphics[width=1\textwidth]{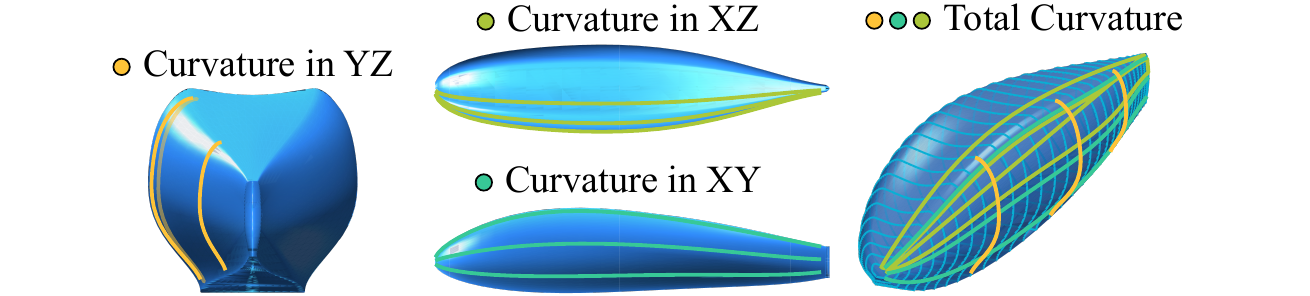}
  	\caption{
  	\textit{Determination of Curvature from Selected Sections.}
  	\newline 3D curvature is estimated by calculating the 2D curvature of representative sections. Along the \textit{yellow} lines we measure curvature in the YZ axis, along the \textit{green} lines in XZ, and along the \textit{blue} lines in XY. The total curvature is the mean curvature of all lines.}
	\label{fig:curvature}
\end{figure}

Rather than precisely measuring the three-dimensional curvature of the millions of designs produced in a single run of SAIL, for the sake of computational efficiency, we estimate the curvature based on a few fixed regions. We calculate the two-dimensional curvature along nine lines (shown in Figure~\ref{fig:curvature}): three in each of the XY, XZ, and YZ dimensions. We take the mean of these curvatures as an estimate of the curvature of the design. We calculate curvature $K$ analytically from each pair of neighboring points along the line as $K =  \frac{\lvert x'y'' - y'x'' \rvert}{(x'^2 + y'^2)^\frac{3}{2}}$ where $x$ and $y$ are their cartesian coordinates.

In high-performing velomobiles, curvature in the shell allows designers to minimize the effects of side winds while simultaneously reducing the frontal area and maintaining enough space to accommodate the rider's feet, knees, and the machinery of the bicycle. The addition of curvature in one part of the design may also require curvature in another, in order to effectively guide and reattach the airflow. In addition to aerodynamic concerns, curvature is introduced into designs to improve the structural integrity of the shell. The shell of the vehicle is thin, and so where the shell is flat it is also weak, weak enough to buckle and change shape at high speed or in high winds. 

The ability to effectively explore features like curvature, which are not directly correlated to performance, but whose effects we are interested in, is a design exploration capability that is difficult to replicate with techniques that rely on producing variety through trade-offs, such as multi-objective optimization.

\subsubsection*{Experimental Setup}	
Properly evaluating three-dimensional designs requires a computationally intensive flow simulator, rather than the purpose built solver as was used for two-dimensional airfoils. The computational expense of using fluid dynamics simulations, however, means that evaluating every design in a prediction map at every step of the algorithm, as we did in the airfoil case, is infeasible.
This computational cost prevents us from tracking the improvement of design performance and modeling accuracy in the detailed way that was possible in the relatively inexpensive airfoil case. 

Comparisons with a black-box optimization algorithm such as CMA-ES also presents some difficulty. Without a simple mapping between parameter and feature space, it is difficult to search within a single bin. The awkwardness of such a mapping aside, the computational cost alone of running CMA-ES several hundred times is prohibitively expensive.
\footnote{Assuming a budget of 1000 PE per run, an estimated 3000 hours (125 days) of computation would be required to run CMA-ES once in each of the 625 bins in a 25 X 25 prediction map, using 80 2.6ghz cores.}

These difficulties make it infeasible to solve this problem with a traditional optimizer, so the comparisons made here are between the designs produced by SAIL using different encodings. In the 2D airfoil test case SAIL was shown to be capable of finding near-optimal designs, and we assume here that it likewise produces designs which are high-performing, even if not optimal. If different encodings produce designs with similar performance and reveal the same feature relationships this consistency will give us confidence in SAIL's performance.

 We initialize SAIL with 200 samples drawn from a Sobol sequence and add 10 samples from the acquisition map at every illumination iteration, for a total of 1000 samples. We divide the feature space into a 25 X 25 acquisition map for a total of 625 bins. Fitness values are approximated using a single GP model with a squared exponential kernel with ARD (see Section \ref{sec:gp}, \textit{Gaussian Process Models}) which predicts drag force on the design. Though fitness values are approximated, the features of each design are derived precisely, and every design is fully constructed.
 All results shown are median values over 20 replicates
 We evaluate the fitness of the produced designs purely on aerodynamic criteria. Velomobile shells are judged by the drag force they produce when traveling at 20 m/s (72km/h). Flow simulation is performed with the OpenFOAM Computational Fluid Dynamics toolbox~\citep{openfoam}.

\section{Encoding Comparison}
	\subsubsection*{Design Performance}

\begin{figure}[h!]
	\centering
	\includegraphics{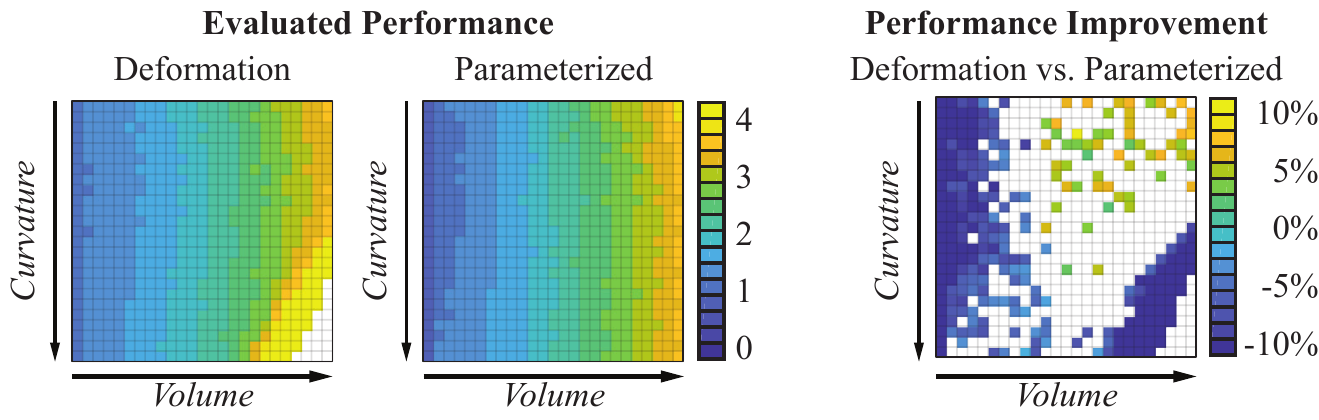}

  	\caption{\textit{True Performance of Designs Produced by SAIL with Different Representations.}\newline The resulting fitness when every design in the prediction map is evaluated in the simulator (fitness is drag force in Newtons, lower is better). Prediction maps created after running SAIL with a budget of 1000 precise evaluations for each encoding, with each bin containing the median fitness found over 20 replicates. Only significant comparisons ($p < 0.05$ determined by Mann-Whitney-Wilcoxon test) are shown, with lighter colors indicating better performance by the deformation encoding. Both encodings capture the strong relationship between volume and drag, but excel in different feature regions, with the deformation performing better in low-curvature, large-volume regions, but unable to express high-volume, high-curvature designs.}
	\label{fig:fitmaps}
\end{figure}

The performance of the designs produced through SAIL (Figure~\ref{fig:fitmaps}) illustrates the strengths of each encoding. While the free-form deformation is able to produce higher performing solutions in the high-volume low-curvature regions of the feature space, it is not flexible enough, or its parameter ranges are too limited, to produce designs which have both very high volume and high curvature. 

Comparison of performance maps reveals that the trend of high-curvature high-volume shapes performing poorly compared to low-curvature high-volume shapes is a quirk in the deformation encoding, not an underlying relationship. Few FFD solutions can create this level of curvature in the high-volume case, and are drawn from a small pool of possible designs with this combination of features. In the feature/performance maps created with the parameterized representation, volume clearly dominates curvature, and high-curvature designs are at no clear disadvantage. 

The feature region where the FFD encoding is unable to produce designs is an exception that highlights the similarity of performance of the encodings in the other regions. In more than two-thirds of bins the fitnesses of the designs found by the two encodings is within $5\%$, with those greater than $5\%$ found at the edges of the feature ranges. This similarity demonstrates the capability of SAIL to illuminate the relationship between features and performance, independent of the representative power of the particular encoding. 


	\subsubsection*{Model Accuracy}
Modeling performance in this more realistic problem is more difficult for several reasons: the encodings have more degrees of freedom (16 dimensions as opposed to 10 in the 2D airfoil experiments), the parameters of the encodings are not as closely linked to performance (as in 2D PARSEC airfoils), and the problem itself is much more complicated (three-dimensional versus two-dimensional flow). While we are not trying to predict the performance of any and all designs, we are still targeting a much larger set of solutions than is typical of surrogate-assisted optimization methods, which seek only to model performance around the global optimum. 

Though the rank-based optimization process of MAP-Elites is forgiving of errors, and the bar for optimality in design exploration is lower than for pure optimization, to have confidence in our results we must first be confident that our models can predict performance with a degree of accuracy. At the end of each run of SAIL a prediction map was produced and every design evaluated in our flow simulator. 

\begin{figure}[ht]
	\centering
	\includegraphics{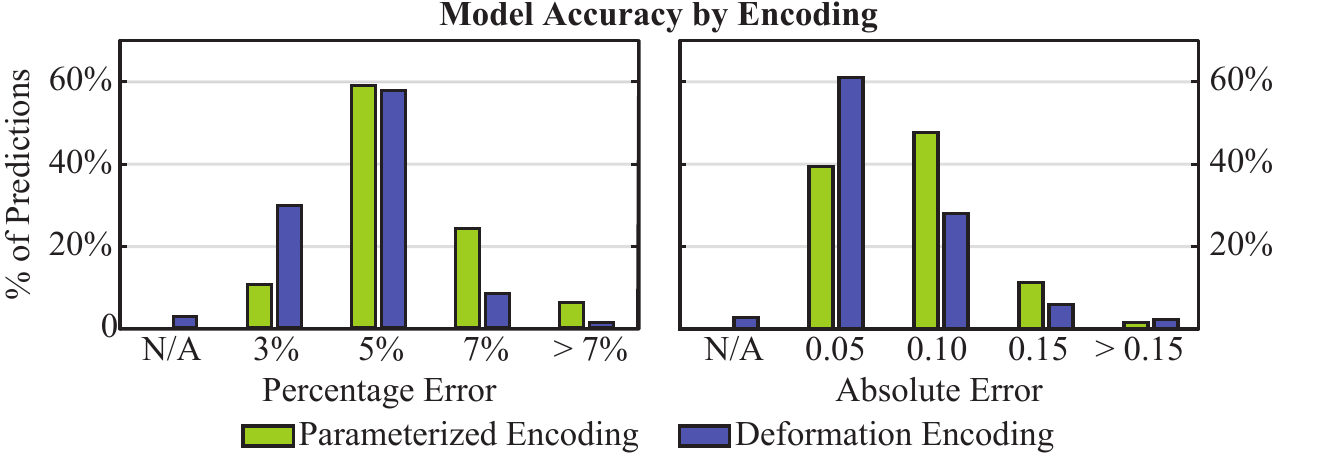}
  	\caption{\textit{Model Accuracy in Prediction Map by Encoding.} \newline
  	Distribution of prediction error of designs in the prediction maps produced after 1000 precise evaluations (median values over 20 replicates). In the case of the deformation encoding not all bins were explored, but are still counted toward the total so as not to distort the comparison between the encodings. Fitness values of designs ranged between 0.8 N and 5.8 N. 
  	} 
	\label{fig:accuracy}
\end{figure}

For both encodings the performance of the majority of designs was predicted within 5\% of the values found in simulation (Figure~\ref{fig:accuracy}, left). Though the FFD encoding was not able to produce designs in every feature region, overall the performance of the designs produced was easier to predict. The predicted performance of the majority of designs created using the deformation encoding was accurate within 0.05 N, while the majority of designs created using the parameterized encoding were only accurate within 0.10 N (Figure~\ref{fig:accuracy}, right). 

	\subsubsection*{Design Exploration}
  
\begin{figure}[h]
	\centering
	\includegraphics{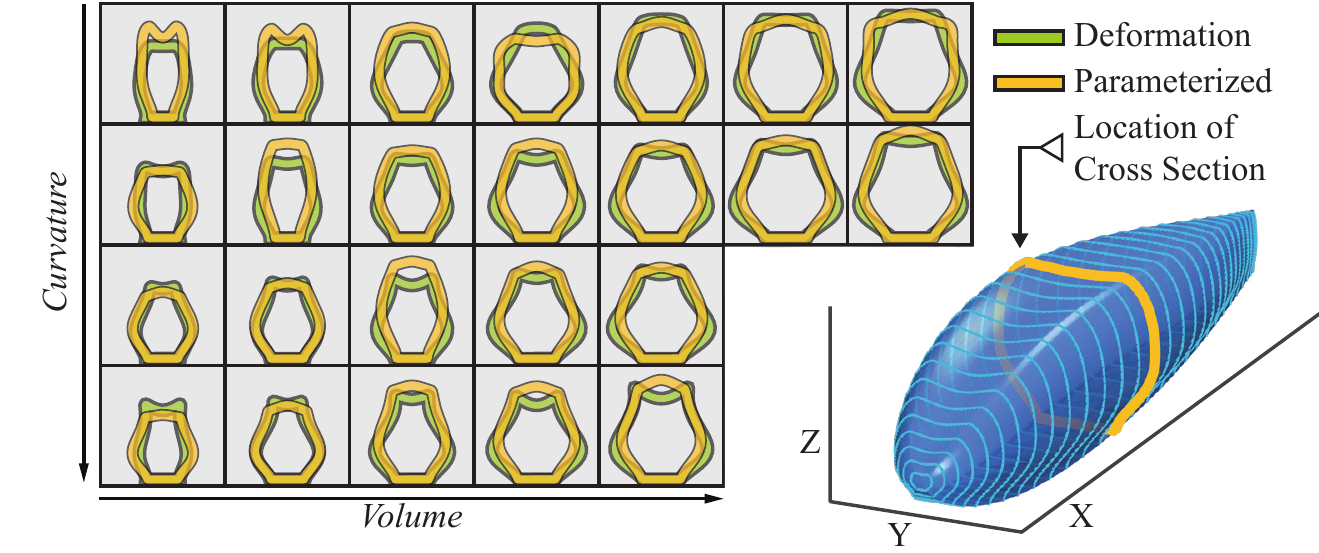}
  	\caption{\textit{Cross-Sections of Optimal Designs in Each Feature Region by Encoding.}\newline 
  	Middle cross section of optimal designs in selected feature regions, free form deformation in green, parameterized encoding in yellow. The different degrees of freedom to each encoding produce differing solutions for the same feature combinations. 
    }
	\label{fig:crossSection}
\end{figure}

Different encodings may lead to different solutions to the same problem, but SAIL is able to find diverse, high-performing examples and accurately predict their performance regardless. Though a variety of solutions is produced by both the parameterized and deformation representations, the designs produced by each tend towards different themes. By examining the cross-sections of designs from each encoding in the same feature region, these biases are revealed (Figure~\ref{fig:crossSection}). 

\begin{figure}[h!]
  \centering
  \includegraphics{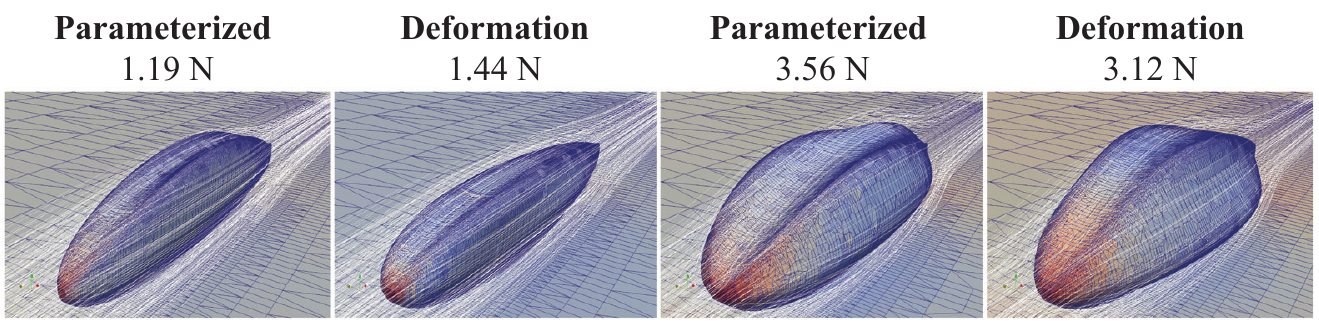}
    \caption{\textit{Different Representations Lead to Different Design Approaches.}\newline 
    Sample low-curvature designs produced by the parameterized and FFD encodings. Left: low-volume, Right: high-volume. Meshes colored by air pressure, with warmer colors indicating higher pressures. The parameterized encoding minimizes frontal area leading to a taller, thinner designs, reducing drag in the low-volume case. In the high-volume case the smoother designs created by FFD achieve higher performance.}
  \label{fig:flow}
\end{figure}

The designs produced by the parameterized encoding are typically taller and thinner designs, lowering drag by reducing the frontal area which first hits the air. Designs produced by FFD have less flexibility at the leading edge, and earn higher fitness with smoother designs that guide flow from a larger frontal area. These strategies are shown in Figure~\ref{fig:flow} for high- and low-volume cases.

\begin{figure}[h!]
  \centering
  \includegraphics{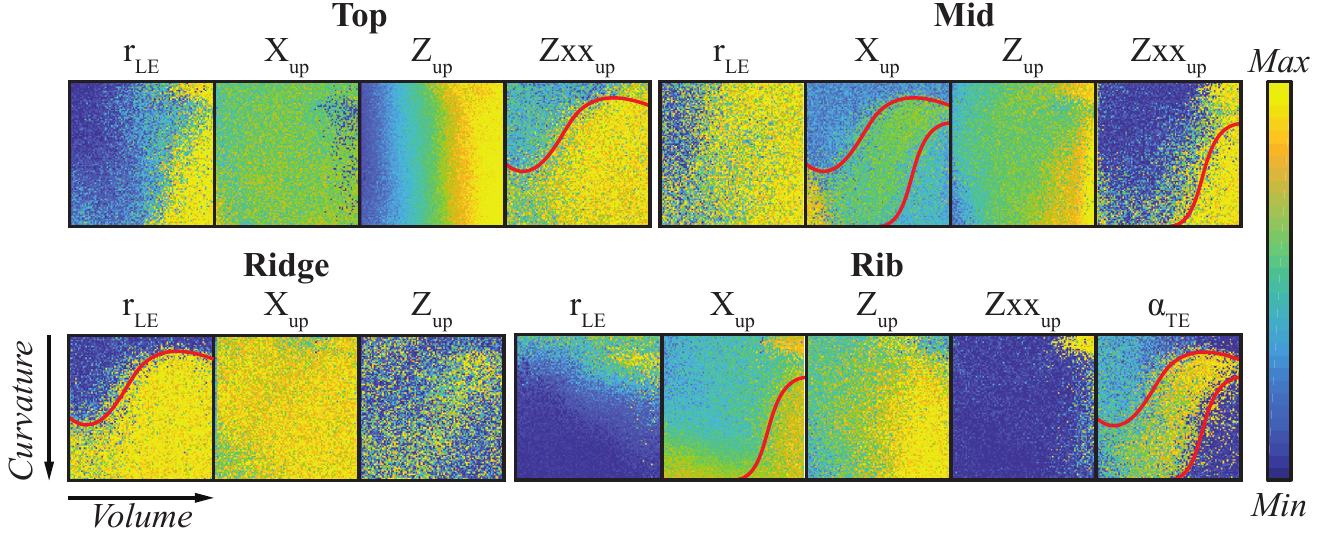}
    \caption{\textit{Visualizing Parameter Values of Optimal Designs}\newline
    Parameter values of optimal parameterized designs in a $250\times250$ bin prediction map. Lighter colors indicate a value closer to the top of the parameter range. 
    Visualizing the parameter ranges used in the optimal designs can aid designers in setting parameter bounds in an encoding. 
    At a glance it is apparent for some parameters, such as the Ridge $X_{up}$ of the Rib $Zxx_{up}$, the range should be shifted, as the optimal values are all at one extreme. The optimal values of the Top $X_{up}$ are all mid range values, hinting that this range could be tightened for greater precision, while the noisy values of the Ridge $Z_{up}$ may indicate that the range is too small. 
    Visualizing the parameters of the optimal designs allows us to easily spot parameter relationships and design regions, such as the correlated design region borders in red, where sudden transitions in optimal values occur across multiple parameters in the same feature regions.
    }
  \label{fig:parameters}
\end{figure}

While general design concepts can be discerned by browsing optimal designs, a more detailed understanding can be gained by viewing the parameters values of the optimal designs through a feature space lens. Figure~\ref{fig:parameters} shows 16 maps, one for each variable of the parameterized encoding. Each map is colored according to the value of the respective parameter, showing the 16 values behind each design in a typical prediction map.

Visualizing the parameter values of the large number of designs produced by SAIL can allow designers to understand the interaction of parameters and features, and to tune their encodings by removing or introducing additional degrees of freedom or adjusting the range of existing parameters. When, for example, the optimal values are all at the edge of a parameter range, that range can be extended or shifted; when the optimal values for a parameter are noisy then it follows that the value has little effect on fitness, meaning it is has no effect, or either unnecessary or the range is too narrow. 

Correlations between parameters can also be easily detected, even visually. 
These correlations could be clues to underlying design concepts, allowing high-performing design prototypes to be identified~\citep{Hagg2018}. If parameters are correlated across the entire feature space they may be candidates for collapsing into a single degree of freedom, reducing the dimensionality of the problem and predisposing the encoding to faster convergence. With a large set of high-performing solutions statistical techniques, such as analysis of variance, could also be applied to analyze new representations in a way that would be impossible with only a handful of designs.



\section{Computation Cost}

	Whenever fitness approximation techniques are applied, the trade-off between complexity of modeling and the expense of precise evaluation must be considered. In the case of illumination, because of the large number of evaluations required, it is not difficult to tilt the balance in favor of modeling.

\begin{figure}[h!]
	\centering
	\includegraphics{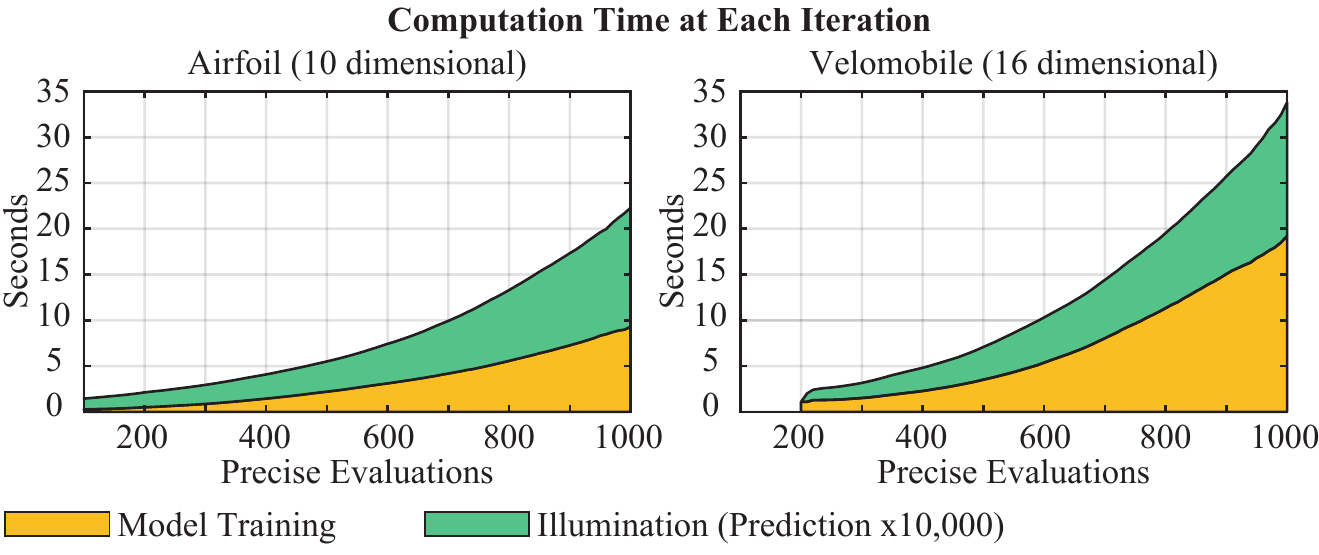}
  	\caption{\textit{Model Training and Prediction Costs in Each Domain.}\newline
  	Model training and prediction time over the course of a typical run, with 1 core of a 3.1 GHz processor (moving average). Values denote the time taken in a single iteration for computation costs of training a single GP model (\textit{yellow}) and the time required to perform the 10,000 predictions used to create a single acquisition map (\textit{green}).
  	}
	\label{fig:timingItr}
	\end{figure}

	The cost of training and prediction with Gaussian process models can, however, become significant when larger data sets are considered. When applying GP models for exact inference, complexity is cubic in the size of the data set. Though starting from a small base cost, exact inference becomes infeasible when more than a few thousand samples are considered. This trend of increasing complexity can be seen in our own experiments (Figure \ref{fig:timingItr}), where the cost of training and prediction increases by nearly ten times from the first iteration to the last. 

	SAIL was designed with expensive domains and small data sets in mind, and where data is more plentiful more sophisticated modeling techniques could be used to maintain performance. There are several methods which allow GPs to cope with larger data sets. Data can, for instance, be partitioned into separate groups~\citep{snelson2007local}, or lower rank approximations can be made of the covariance matrix based around representative ``inducing points''~\citep{quinonero2005unifying, quia2010sparse}.

\begin{figure}[h!]
	\centering
	\includegraphics{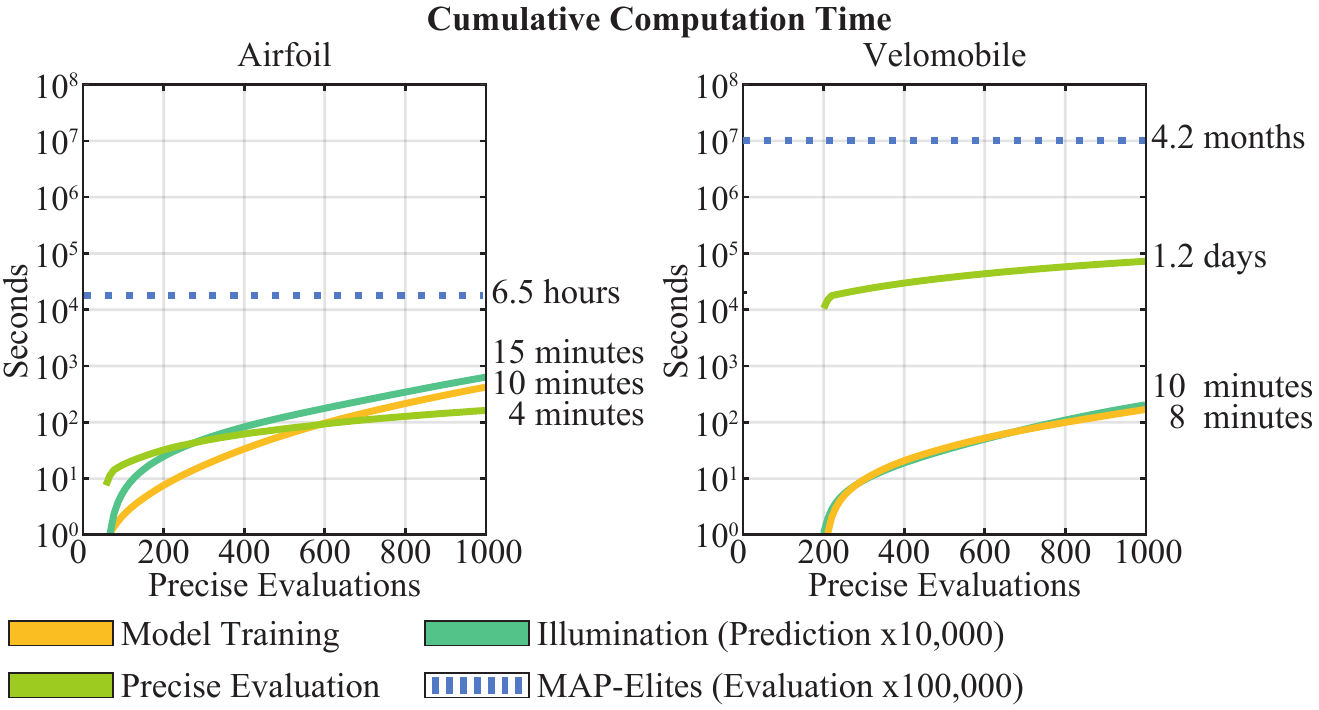}
  	\caption{\textit{Cumulative Computational Cost of SAIL Components.}\newline
  	Cumulative cost of each SAIL component in a typical run, using 1 core of a 3.1 GHz processor. Time in logarithmic scale. One illumination consists of 10,000 predictions. Model training and prediction time in the airfoil case includes both the drag and the lift models. Model training and Illumination are each performed 96 times in the airfoil case and 81 times in the velomobile case. 1000 precise evaluations are performed in both cases. For comparison the approximate computational cost of a 100,000 precise evaluation MAP-Elites run is also shown.
  	\newline
  	}
	\label{fig:timingCumul}
	\end{figure}

	Though the computational cost of modeling and prediction increases cubically, we find that SAIL is still more efficient than using precise evaluations alone, even in inexpensive cases. In Figure \ref{fig:timingCumul} we examine the cumulative computational costs of model training, prediction, and precise evaluation. In the airfoil case one precise evaluation requires only a fraction of a second, and more time is spent on model training and prediction than on precise evaluation. At each iteration SAIL performs 10,000 predictions to illuminate an acquisition map, with a total of nearly 1 million predictions over a single run. In our experiments, MAP-Elites was given a budget of 100,000 precise evaluations, or 6.5 hours of evaluation time. Even with this additional computation and exact results, the solutions found were still much worse than those found using 30 minutes of SAIL's combination of evaluation and modeling (see Figure~\ref{fig:progress} in Section~\ref{sec:2d}).

If the airfoil case illustrates the potential for SAIL to accelerate illumination even in inexpensive domains, the velomobile case demonstrates how SAIL can extend the reach of quality-diversity algorithms. In this more expensive case, SAIL spends more than a day of computation time evaluating 1000 designs and less than 20 minutes on model building and prediction. If we were to attempt to run MAP-Elites in this domain, the 100,000 evaluations would require more than 4 months of computation time. The illumination of the design space in this domain is \textit{only} possible because of the surrogate modeling techniques in SAIL.

\section{Discussion}

The SAIL algorithm produces a model of the objective function in high-fitness regions across the feature space, even with a limited computational budget. With the aid of these models, illumination algorithms can produce a diversity of high-performing designs which reveal relationships between features as well as biases in encodings. 

Our experiments have shown that SAIL is effective without a carefully tuned domain specific encoding, which opens up the possibility of using it as a tool to assist in the creation and tuning of new encodings. Whether testing the capabilities of a newly designed representation, or iteratively improving an existing encoding, SAIL provides a way of understanding the inherent biases of a representation. Even if the encoding is destined for use in a more traditional optimization algorithm, SAIL allows a designer to visualize the variety of designs an encoding is able to express, the optima they are able to reach, and the ease of modeling their performance from their genotype. SAIL does not require powerful encodings, and could be used as a tool to create them.

In design exploration, the chief advantage illumination approaches have over multi-objective approaches is the ability to explore features which are not in opposition, or that have no relation to fitness at all. These features could be those which have an unknown effect on performance which designers would like to better understand, such as the percentage of a structure created with a new material, or features whose goal is not to improve performance, such as aesthetics.

Multi-objective optimization is used to power automated design exploration approaches known as `innovization'~\citep{deb2006innovization}. These approaches use optimization algorithms to reveal new design principles by searching for commonalities in sets of high-performing designs. 
The ability to produce designs in larger variety, and which vary across more dimensions of freedom, make illumination techniques an ideal fit to produce the required raw material of diverse high-performing solutions.

That prediction maps are most intuitively visualized in two dimensions does not limit the use of SAIL to only a pair of features. Prediction maps are created using a continuous model built from samples selected from the acquisition map. The acquisition map itself is merely a collection of candidate solutions for the model, and so has no direct connection to the prediction map. As such the form of the prediction map is not bound by the structure of the acquisition map: the resolution of the map could be different, feature ranges could be tuned to zoom in on a particular feature region, feature dimensions themselves could even be added or removed. For example, the 250$\times$250 parameter maps in Figure~\ref{fig:parameters} come from a prediction map produced using a model created by running SAIL with a 25$\times$25 acquisition map. 

To allow for easier visualization, it is possible to run SAIL to build models using acquisition maps with many feature dimensions, and to then produce prediction maps which only examine the relationship between two of the features at a time. As the optimization algorithm which produces the prediction map is based only on the model, prediction maps can be produced with very little computation, even at high resolution. The acceleration provided by surrogate-assistance enables SAIL to not only act as a data-efficient method to perform illumination, but also allows the production of maps which are essentially continuous. 

	It may be necessary or advantageous to estimate the behavior or features of a given solution in addition to their performance. In many cases where a data-efficient version of MAP-Elites would be most useful, such as robotics, the feature or behavior description is obtained during evaluation, e.g. how active each leg of a hexapod is during a gait with a given controller~\citep{cully2015robots}. Even in design cases, where simulation may not be necessary to derive features, instantiating the design from the genome can still represent a comparatively large computational expense, given that it must be done millions of times over the course of the algorithm. In these design cases models which estimate features could further accelerate SAIL, and given these samples ready availability could be made very accurate.

	Integration of additional surrogate models could do more than accelerate.
	In our experiments when a design was simulated but did not converge, whether due to numerical instabilities or odd geometries that a simple solver like XFoil was not designed to model, the result was simply discarded and the next design taken in its place.  If used on truly expensive problems, this approach is insufficient: we cannot afford to simply throw away data. Additional models could be used to estimate the likelihood of a design to converge in simulation, guiding the illumination process towards more robust designs. In evolutionary robotics similar approaches have been proposed to produce controllers which bridge the reality gap, avoiding solutions which are unlikely to work in the real world in the way that they do in simulation~\citep{koos2013transferability}.

Though MAP-Elites has shown remarkable potential, the intensive computation it requires precludes its use in many domains. By pairing MAP-Elites with surrogate assistance, a Bayesian optimization equivalent for illumination is created. By enabling illumination in computationally expensive domains, SAIL opens up new avenues for experiments and applications of quality-diversity techniques, especially in design. 

The capability to rapidly understand the performance potential of the design space through concrete high-performing examples is a potential boon to designers. SAIL not only accelerates the generative design cycle, but allows the effect of user-defined features to be examined, adding new flexibility to cooperative human-machine design exploration. Generative design tools which consider more than objectives, such as SAIL, can help designers explore what is possible, beyond what is optimal.

\section*{Acknowledgments}

This work received funding from the European Research Council (ERC) under the European Union's Horizon 2020 research and innovation programme (grant agreement number 637972, project "ResiBots") and the German Federal Ministry of Education and Research (BMBF) under the Forschung an Fachhochschulen mit Unternehmen programme  (grant agreement number 03FH012PX5, project "Aeromat"). The authors would like to thank Alexander Hagg and the ResiBots team for all their feedback. 

\section*{Source Code}
The source code used to conduct the experiments in this publication is available with an open-source license at: \url{http:// www.github.com/agaier/sail_ecj2018}

\small

\bibliographystyle{apalike}
\bibliography{sailBib_ecj}

\end{document}